# Graphical Nonconvex Optimization for Optimal Estimation in Gaussian Graphical Models

Qiang Sun[*], Kean Ming Tan[†], Han Liu[‡] and Tong Zhang[§]

**Abstract**

We consider the problem of learning high-dimensional Gaussian graphical models. The graphical lasso is one of the most popular methods for estimating Gaussian graphical models. However, it does not achieve the oracle rate of convergence. In this paper, we propose the graphical nonconvex optimization for optimal estimation in Gaussian graphical models, which is then approximated by a sequence of convex programs. Our proposal is computationally tractable and produces an estimator that achieves the oracle rate of convergence. The statistical error introduced by the sequential approximation using the convex programs are clearly demonstrated via a contraction property. The rate of convergence can be further improved using the notion of sparsity pattern. The proposed methodology is then extended to semiparametric graphical models. We show through numerical studies that the proposed estimator outperforms other popular methods for estimating Gaussian graphical models.

**Keywords**: Adaptivity, Graphical nonconvex optimization, Nonconvexity, Semiparametric, Sequential convex approximation.

## 1 Introduction

We consider the problem of learning an undirected graph $G = (V, E)$, where $V = \{1, \ldots, d\}$ contains nodes that represent $d$ random variables, and the edge set $E$ describes the pairwise conditional dependence relationships among the $d$ random variables. Gaussian graphical models have been widely used to represent pairwise conditional dependencies among a set of variables. Let $\boldsymbol{X}$ be a $d$-dimensional random variables. Under the Gaussian assumption $\boldsymbol{X} \sim \mathcal{N}(\mathbf{0}, \boldsymbol{\Sigma}^*)$, the graph $G$ is encoded by the sparse concentration matrix $\boldsymbol{\Omega} = (\boldsymbol{\Sigma}^*)^{-1}$, or the sparse inverse correlation matrix $\boldsymbol{\Psi}^* = (\mathbf{C}^*)^{-1}$. Here, $\mathbf{C}^*$ is the correlation matrix such that $\boldsymbol{\Sigma}^* = \mathbf{W}\mathbf{C}^*\mathbf{W}$ and $\mathbf{W}^2$ is a diagonal matrix with diagonal elements of $\mathbf{\Sigma}^*$. In particular, it is well known that the $j$th and $k$th variables are conditionally independent given all of the other variables if and only if the $(j,k)$-th element of $\mathbf{\Omega}^*$ (or $\mathbf{\Psi}^*$) is equal to zero. Thus, inferring the conditional dependency structure of a Gaussian graphical model boils down to estimating a sparse inverse covariance (or correlation) matrix.

[*]Department of Operations Research and Financial Engineering, Princeton University, Princeton, NJ 08544; e-mail: qiangs@princeton.edu.

[†]Department of Operations Research and Financial Engineering, Princeton University, Princeton, NJ 08544; e-mail: kmtan@princeton.edu.

[‡]Department of Operations Research and Financial Engineering, Princeton University, Princeton, NJ 08544, USA; e-mail:hanliu@princeton.edu.

[§]Tencent AI Lab, Shen Zhen, Guangdong, China; e-mail:tongzhang@tongzhang-ml.org.

A number of methods have been proposed to estimate the sparse concentration matrix under the Gaussian assumption. For example, Meinshausen and Bühlmann (2006) proposed a neighborhood selection approach for estimating Gaussian graphical models by solving a collection of sparse linear regression problems using the lasso penalty. In addition, Yuan (2010) and Cai et al. (2011) proposed the graphical Dantzig and CLIME, both of which can be solved efficiently. From a different perspective, Yuan and Lin (2007) and Friedman et al. (2008) proposed the graphical lasso methodology, a penalized likelihood based approach, to estimate the concentration matrix $\mathbf{\Omega}^*$ directly. Various extensions of the graphical lasso were proposed and the theoretical properties were also studied (among others, Banerjee et al., 2008; Rothman et al., 2008; Ravikumar et al., 2011). The Gaussian graphical models literature is vast and we refer the reader to Cai et al. (2016a) and Drton and Maathuis (2016) for recent reviews on this topic.

Despite the large literature on using the graphical lasso to estimate concentration matrices in Gaussian graphical models, the graphical lasso does not achieve the oracle rate of convergence. More specifically, it is belived that the optimal rate of convergence in spectral norm for the graphical lasso is at the order of $\sqrt{s \log d/n}$ (Rothman et al., 2008). Here, $n$ is the sample size, $d$ is the number of nodes, and $s$ is the number of edges in the true graph. In fact, the graphical lasso and all of the aforementioned methods are based on the lasso penalty and it is well known that convex penalties usually introduce non-negligible estimation bias. For example, in the linear regression setting, Fan and Li (2001); Zhang (2010a,b); Fan et al. (2017) have shown that the nonconvex penalized regression is able to eliminate the estimation bias and attain a more refined statistical rate of convergence.

Based on these insights, we consider the following penalized maximum likelihood estimation with nonconvex regularizers:

$$\widehat{\mathbf{\Theta}} = \operatorname*{argmin}_{\mathbf{\Theta} \in \mathcal{S}_+^d} \left\{ \langle \mathbf{\Theta}, \widehat{\mathbf{\Sigma}} \rangle - \log \det(\mathbf{\Theta}) + \sum_{i \neq j} p_\lambda(\Theta_{ij}) \right\}, \tag{1.1}$$

where $\mathcal{S}_+^d = \{\mathbf{A} \in \mathbb{R}^{d \times d} : \mathbf{A} = \mathbf{A}^{\mathrm{T}}, \mathbf{A} \succ 0\}$ is the symmetric definite cone formed by all symmetric positive definite matrices in $d \times d$ dimensions, $\widehat{\mathbf{\Sigma}}$ is the sample covariance matrix, and $p_\lambda(\cdot)$ is a nonconvex penalty. Here, $\langle \mathbf{A}, \mathbf{B} \rangle = \mathrm{tr}(\mathbf{A}^{\mathrm{T}}\mathbf{B})$ denotes the trace of $\mathbf{A}^{\mathrm{T}}\mathbf{B}$. However, from the computational perspective, minimizing a folded concave penalized problem is very complicated due to its intrinsic nonconvex structure. Indeed, Ge et al. (2015) have shown that solving (1.1) with a general concave penalty, such as the SCAD Fan and Li (2001) or the MCP Zhang (2010a), is strongly NP-hard. In other words, there does not exist a fully polynomial-time approximation scheme for problem (1.1) unless more structures are assumed. Recently, Loh and Wainwright (2015) proposed an algorithm to obtain a good local optimum for (1.1), but an additional convex constraint that depends on the unknown true concentration matrix is imposed. Moreover, they failed to provide a faster rate of convergence statistically due to not taking the signal strength into account.

In this paper, instead of directly solving the nonconvex problem (1.1), we propose to approximate it by a sequence of adaptive convex programs. Even though the proposed approach is solving a sequence of convex programs, under some regularity conditions, we show that the proposed estimator for estimating the sparse concentration matrix achieves the oracle rate of convergence of $\sqrt{s/n}$, treating as if the locations of the nonzeros were known a priori. This is achieved by a contraction property. Roughly speaking, each convex program gradually contracts the initial estimator to the region of oracle rate of convergence even when a bad initial estimator is used in the first place:

$$\|\widehat{\boldsymbol{\Psi}}^{(\ell)} - \boldsymbol{\Psi}^*\|_{\mathrm{F}} \leq \underbrace{C\sqrt{\frac{s}{n}}}_{\text{Oracle Rate}} + \underbrace{\frac{1}{2}\|\widehat{\boldsymbol{\Psi}}^{(\ell-1)} - \boldsymbol{\Psi}^*\|_{\mathrm{F}}}_{\text{Contraction}},$$

where $\widehat{\boldsymbol{\Psi}}^{(\ell)}$ is the inverse correlation matrix estimator after the $\ell$-th convex approximation, $\|\cdot\|_{\mathrm{F}}$ denotes the Frobenius norm, $C$ is a constant, and $\sqrt{s/n}$ is referred to as the oracle rate. Each iteration of the proposed method helps improve the accuracy only when $\|\widehat{\boldsymbol{\Psi}}^{(\ell-1)} - \boldsymbol{\Psi}^*\|_{\mathrm{F}}$ dominates the statistical error. The error caused by each iteration is clearly demonstrated via the proven contraction property. By rescaling the inverse correlation matrix using the estimated marginal variances, we obtain an estimator of the concentration matrix with spectral norm convergence rate in the order of $\sqrt{\log d/n} \vee \sqrt{s/n}$. Here, $a \vee b = \max\{a, b\}$ is used to denote the maximum of $a$ and $b$. By exploiting a novel notion called sparsity pattern, we further sharpens the rate of convergence under the spectral norm.

The rest of this paper proceeds as follows. In Section 2, we propose the new methodology and its implementation. Section 3 is devoted to theoretical studies. We show that the proposed methodology can be extended to the semiparametric graphical models in Section 4. Numerical experiments are provided to support the proposed methodology in Section 5. We conclude the paper in Section 6. All the proofs and technical details are collected in the supplementary material.

**Notation**: We summarize the notation that will be used regularly throughout the paper. Given a vector $\mathbf{u} = (u_1, u_2, \ldots, u_d)^{\mathrm{T}} \in \mathbb{R}^d$, we define the $\ell_q$-norm of $\mathbf{u}$ by $\|\mathbf{u}\|_q = (\sum_{j=1}^d |u_j|^q)^{1/q}$, where $q \in [1, \infty)$. For a set $\mathcal{A}$, let $|\mathcal{A}|$ denote its cardinality. For a matrix $\mathbf{A} = (a_{i,j}) \in \mathbb{R}^{d\times d}$, we use $\mathbf{A} \succ 0$ to indicate that $\mathbf{A}$ is positive definite. For $q \geq 1$, we use $\|\mathbf{A}\|_q = \max_{\mathbf{u}} \|\mathbf{A}\mathbf{u}\|_q/\|\mathbf{u}\|_q$ to denote the operator norm of $\mathbf{A}$. For index sets $\mathcal{I}, \mathcal{J} \subseteq \{1, \ldots, d\}$, we define $\mathbf{A}_{\mathcal{I},\mathcal{J}} \in \mathbb{R}^{d\times d}$ to be the matrix whose $(i, j)$-th entry is equal to $a_{i,j}$ if $i \in \mathcal{I}$ and $j \in \mathcal{J}$, and zero otherwise. We use $\mathbf{A} \odot \mathbf{B} = (a_{ij}b_{ij})$ to denote the Hadamard product of two matrices $\mathbf{A}$ and $\mathbf{B}$. Let $\mathrm{diag}(\mathbf{A})$ denote the diagonal matrix consisting diagonal elements of $\mathbf{A}$. We use $\mathrm{sign}(x)$ to denote the sign of $x$: $\mathrm{sign}(x) = x/|x|$ if $x \neq 0$ and $\mathrm{sign}(x) = 0$ otherwise. For two scalars $f_n$ and $g_n$, we use $f_n \gtrsim g_n$ to denote the case that $f_n \geq cg_n$, and $f_n \lesssim g_n$ if $f_n \leq Cg_n$, for two positive constants $c$ and $C$. We say $f_n \asymp g_n$, if $f_n \gtrsim g_n$ and $f_n \lesssim g_n$. $\mathcal{O}_{\mathbb{P}}(\cdot)$ is used to denote bounded in probability. We use $c$ and $C$ to denote constants that may vary from line to line.

## 2 A Sequential Convex Approximation

Let $\boldsymbol{X} = (X_1, X_2, \ldots, X_d)^{\mathrm{T}}$ be a zero mean $d$-dimensional Gaussian random vector. Then its density can be parameterized by the concentration matrix $\boldsymbol{\Theta}^*$ or the inverse correlation matrix $\boldsymbol{\Psi}^*$. The family of Gaussian distributions respects the edge structure of a graph $G = (V, E)$ in the sense that $\Psi^*_{ij} = 0$ if and only if $(i, j) \notin E$. This family is known as the Gauss-Markov random field with respect to the graph $G$. The problem of estimating the edge corresponds to parameter estimation, while the problem of identifying the edge set, i.e., the set $E \equiv \{i, j \in V \mid i \neq j, \Psi^*_{ij} \neq 0\}$, corresponds to the problem of model selection.

Given $n$ independent and identically distributed observations $\{\boldsymbol{X}^{(i)}\}_{i=1}^n$ of a zero mean $d$-dimensional random vector $\boldsymbol{X} \in \mathbb{R}^d$, we are interested in estimating the inverse correlation matrix $\boldsymbol{\Psi}^*$ and concentration matrix $\boldsymbol{\Theta}^*$. Let $\widehat{\boldsymbol{\Sigma}} = n^{-1} \sum_{1 \le i \le n} \boldsymbol{X}^{(i)} (\boldsymbol{X}^{(i)})^{\mathrm{T}}$ be the sample covariance matrix and let $\widehat{\mathbf{C}} = \widehat{\mathbf{W}}^{-1} \widehat{\boldsymbol{\Sigma}} \widehat{\mathbf{W}}^{-1}$, where $\widehat{\mathbf{W}}^2 = \mathrm{diag}(\widehat{\boldsymbol{\Sigma}})$. To estimate $\boldsymbol{\Psi}^*$, we propose to adaptively solve the following sequence of convex programs

$$\widehat{\boldsymbol{\Psi}}^{(\ell)} = \underset{\boldsymbol{\Psi} \in \mathcal{S}_+^d}{\mathrm{argmin}} \Big\{ \big\langle \boldsymbol{\Psi}, \widehat{\mathbf{C}} \big\rangle - \log\det(\boldsymbol{\Psi}) + \|\boldsymbol{\lambda}^{(\ell-1)} \odot \boldsymbol{\Psi}\|_{1,\mathrm{off}} \Big\}, \quad \text{for} \quad \ell = 1, \ldots, T, \tag{2.1}$$

where $\|\boldsymbol{\Theta}\|_{1,\mathrm{off}} = \sum_{i \neq j} |\Theta_{ij}|$, $\boldsymbol{\lambda}^{(\ell-1)} = \lambda \cdot \mathrm{w}\big(\widehat{\Psi}_{ij}^{(\ell-1)}\big)$ is a $d \times d$ adaptive regularization matrix for a given tuning parameter $\lambda$ and a weight function $\mathrm{w}(\cdot)$, and $T$ indicates the number of total convex programs needed. The weight function $\mathrm{w}(\cdot)$ can be taken to be $\mathrm{w}(t) = p'_\lambda(t)/\lambda$, where $p_\lambda(t)$ is a folded concave penalty such as the SCAD or the MCP proposed by Fan and Li (2001) and Zhang (2010a), respectively.

To obtain an estimate for the concentration matrix estimator $\boldsymbol{\Psi}^*$, we rescale $\widehat{\boldsymbol{\Psi}}^{(T)}$ back to $\widetilde{\boldsymbol{\Theta}}^{(T)} = \widehat{\mathbf{W}}^{-1} \widehat{\boldsymbol{\Psi}}^{(T)} \widehat{\mathbf{W}}^{-1}$ after the $T$-th convex program. This rescaling helps improve the rate of convergence for $\widetilde{\boldsymbol{\Theta}}^{(T)}$ significantly by eliminating the effect introduced through the unpenalized diagonal terms. The detailed routine is summarized in Algorithm 1.

**Algorithm 1** A sequential convex approximation for the graphical nonconvex optimization.

**Input**: Sample covariance matrix $\widehat{\boldsymbol{\Sigma}}$, regularization parameter $\lambda$.

**Step 1**: Obtain sample correlation matrix $\widehat{\mathbf{C}}$ by $\widehat{\mathbf{C}} = \widehat{\mathbf{W}}^{-1} \widehat{\boldsymbol{\Sigma}} \widehat{\mathbf{W}}^{-1}$, where $\widehat{\mathbf{W}}^2$ is a diagonal matrix with diagonal elements of $\widehat{\boldsymbol{\Sigma}}$.

**Step 2**: Solve a sequence of graphical lasso problem adaptively

$$\widehat{\boldsymbol{\Psi}}^{(\ell)} = \underset{\boldsymbol{\Psi} \in \mathcal{S}_+^d}{\mathrm{argmin}} \Big\{ \langle \boldsymbol{\Psi}, \widehat{\mathbf{C}} \rangle - \log\det(\boldsymbol{\Psi}) + \|\boldsymbol{\lambda}^{(\ell-1)} \odot \boldsymbol{\Psi}\|_{1,\mathrm{off}} \Big\},$$

$$\text{and } \boldsymbol{\lambda}^{(\ell)} = \lambda \cdot \mathrm{w}(\widehat{\Psi}_{ij}^{(\ell)}), \text{ for } \ell = 1, \ldots, T.$$

**Step 3**: Obtain an estimate of $\boldsymbol{\Theta}^*$ by $\widetilde{\boldsymbol{\Theta}}^{(T)} = \widehat{\mathbf{W}}^{-1} \widehat{\boldsymbol{\Psi}}^{(T)} \widehat{\mathbf{W}}^{-1}$.

The complexity of Step 2 in Algorithm 1 is $O(d^3)$ per iteration: this is the complexity of the algorithm for solving the graphical lasso problem. We will show in the latter section that the number of iteration can be chosen to be $T \approx \log\log d$ based on our theoretical analysis. Algorithm 1 can be implemented using existing `R` packages such as `glasso`.

# 3 Theoretical Results

In this section, we study the theoretical properties of the proposed estimator. We start with the assumptions needed for our theoretical analysis.

## 3.1 Assumptions

Let $S = \{(i,j) : \Theta^*_{ij} \neq 0, i \neq j\}$ be the support set of the off-diagonal elements in $\mathbf{\Theta}^*$. Thus, $S$ is also the support set of the off-diagonal elements in $\mathbf{\Psi}^*$. The first assumption we need concerns the structure of the true concentration and covariance matrices.

**Assumption 3.1** (Structural Assumption)**.** We assume that $|S| \leq s, \|\mathbf{\Sigma}^*\|_\infty \leq M < \infty$, $0 < \varepsilon_1 \leq \sigma_{\min} \leq \sigma_{\max} \leq 1/\varepsilon_1 < \infty$, $0 < \varepsilon_2 \leq \lambda_{\min}(\mathbf{\Theta}^*) \leq \lambda_{\max}(\mathbf{\Theta}^*) \leq 1/\varepsilon_2 < \infty$. Here, $\sigma^2_{\max} = \max_j \Sigma^*_{jj}$ and $\sigma^2_{\min} = \min_j \Sigma^*_{jj}$, where $\mathbf{\Sigma}^* = \big(\Sigma^*_{ij}\big)$.

Assumption 3.1 is standard in the existing literature for Gaussian graphical models (see, for instance, Meinshausen and Bühlmann, 2006; Yuan, 2010; Cai et al., 2016b; Yuan and Lin, 2007; Ravikumar et al., 2011). We need $\sigma_{\min}$ and $\sigma_{\max}$ to be bounded from above and below to guarantee reasonable performance of the concentration matrix estimator (Rothman et al., 2008). Throughout this section, we treat $M, \varepsilon_1, \varepsilon_2$ as constants to simplify the presentation.

The second assumption we need in our analysis concerns the weight functions, which are used to adaptively update the regularizers in Step 2 of Algorithm 1. Define the following class of weight functions:

$$\mathcal{W} = \Big\{ \mathrm{w}(t) : \mathrm{w}(t) \text{ is nonincreasing },\ 0 \leq \mathrm{w}(t) \leq 1 \text{ if } t \geq 0, \mathrm{w}(t) = 1 \text{ if } t \leq 0 \Big\}. \tag{3.1}$$

**Assumption 3.2** (Weight Function)**.** There exists an $\alpha$ such that the weight function $\mathrm{w}(\cdot) \in \mathcal{W}$ satisfies $\mathrm{w}(\alpha\lambda) = 0$ and $\mathrm{w}(u) \geq 1/2$, where $u = c\lambda$ for some constant $c$.

The above assumption on the weight functions can be easily satisfied. For example, it can be satisfied by simply taking $\mathrm{w}(t) = p'_\lambda(t)/\lambda$, where $p_\lambda(t)$ is a folded concave penalty such as the SCAD or the MCP (Fan and Li, 2001; Zhang, 2010a). Next, we impose an assumption on the magnitude of the nonzero off-diagonal entries in the inverse correlation matrix $\mathbf{\Psi}^*$.

**Assumption 3.3** (Minimal Signal Strength)**.** Recall that $S$ is the true support set. The minimal signal satisfies that $\min_{(i,j)\in S} \Psi^*_{ij} \geq (\alpha + c)\lambda \gtrsim \lambda$, where $c > 0$ is the same constant that appears in Assumption 3.2.

Assumption 3.3 is rather mild. In the sub-Gaussian design case, $\lambda$ can be taken to be the order of $\sqrt{\log d/n}$, which diminishes quickly as $n$ increases. It is an analogue to the minimal signal strength assumption frequently assumed in nonconvex penalized regression problems (Fan and Li, 2001; Zhang, 2010a). Taking the signal strength into account, we can then obtain the oracle rate of convergence.

### 3.2 Main Theory

We now present several main theorems concerning the rates of convergence of the proposed estimator for the sparse inverse correlation and the concentration matrices. The following theorem concerns the rate of convergence for the one-step estimator $\widehat{\mathbf{\Psi}}^{(1)}$ obtained from Algorithm 1 when $\ell = 1$.

**Proposition 3.4** (One-step Estimator)**.** Let $\lambda \asymp \sqrt{\log d/n}$. Under Assumption 3.1, we have

$$\big\|\widehat{\mathbf{\Psi}}^{(1)} - \mathbf{\Psi}^*\big\|_{\mathrm{F}} \lesssim \sqrt{\frac{s\log d}{n}}$$

with probability at least $1 - 8/d$,

*Proof of Proposition 3.4.* We collect the proof of Proposition 3.4 in Appendix A in the supplementary material. □

The above proposition indicates that the statistical error under the Frobenius norm for the one-step estimator is at the order of $\sqrt{s\log d/n}$, which is believed to be unimprovable when one-step convex regularization is used (Rothman et al., 2008; Ravikumar et al., 2011). However, when a sequence of convex programs is used as in our proposal, the rate of convergence can be improved significantly. This is demonstrated in the following theorem.

**Theorem 3.5** (Contraction Property)**.** Suppose that $n \gtrsim s\log d$ and take $\lambda$ such that $\lambda \asymp \sqrt{\log d/n}$. Under Assumptions 3.1, 3.2 and 3.3, $\widehat{\mathbf{\Psi}}^{(\ell)}$ satisfies the following contraction property:

$$\big\|\widehat{\mathbf{\Psi}}^{(\ell)} - \mathbf{\Psi}^*\big\|_{\mathrm{F}} \le \underbrace{8\|\mathbf{\Psi}^*\|_2^2\|\nabla\mathcal{L}(\mathbf{\Psi}^*)_S\|_{\mathrm{F}}}_{\text{Oracle Rate}} + \underbrace{\frac{1}{2}\big\|\widehat{\mathbf{\Psi}}^{(\ell-1)} - \mathbf{\Psi}^*\big\|_{\mathrm{F}}}_{\text{Contraction}}, \quad 1 \le \ell \le T,$$

with probability at least $1 - 8/d$. Moreover, if $T \gtrsim \log(\lambda\sqrt{n}) \gtrsim \log\log d$, we have

$$\big\|\widehat{\mathbf{\Psi}}^{(T)} - \mathbf{\Psi}^*\big\|_{\mathrm{F}} = \mathcal{O}_{\mathbb{P}}\bigg(\sqrt{\frac{s}{n}}\bigg).$$

*Proof of Theorem 3.5.* The proof is collected in Appendix A in the supplementary material. □

Theorem 3.5 establishes a contraction property: each convex approximation contracts the initial estimator towards the true sparse inverse correlation matrix until it reaches the oracle rate of convergence: $\sqrt{s/n}$. To achieve the oracle rate, we need to solve no more than approximately $\log\log d$ convex programs. Note that $\log\log d$ grows very slowly as $d$ increases and thus, in practice, we only need to solve a few convex programs to get a better estimator than existing method such as the graphical lasso. The rate of convergence $\sqrt{s/n}$ is better than the existing literature on likelihood-based methods for estimating sparse inverse correlation matrices (Rothman et al., 2008; Lam and Fan, 2009a; Ravikumar et al., 2011). By rescaling, we obtain a concentration matrix estimator with a faster rate of convergence.

**Theorem 3.6** (Faster Rate in Spectral Norm)**.** Under the same conditions in Theorem 3.5, we have

$$\big\|\widetilde{\mathbf{\Theta}}^{(T)} - \mathbf{\Theta}^*\big\|_2 = \mathcal{O}_{\mathbb{P}}\bigg(\sqrt{\frac{s}{n}} \vee \sqrt{\frac{\log d}{n}}\bigg).$$

*Proof of Theorem 3.6.* The proof is deferred to Appendix A in the supplementary material. □

The theorem above provides the optimal statistical rate for estimating sparse concentration matrices using likelihood based methods (Rothman et al., 2008; Lam and Fan, 2009b; Ravikumar et al., 2011). The extra $\log d$ term is a consequence of estimating the marginal variances. We further sharpen the obtained theory using a novel notion, called sparsity pattern, as defined below.

**Definition 3.7** (Sparsity Pattern)**.** For a matrix $\mathbf{A} = (a_{ij})$, we say $\mathbf{A}_{\mathrm{sp}} = (a_{ij}^{\mathrm{sp}})$ is the corresponding sparsity pattern matrix if $a_{ij}^{\mathrm{sp}} = 1$ when $a_{ij} \neq 0$; and $a_{ij}^{\mathrm{sp}} = 0$, otherwise.

Let $\mathbf{M}^*$ be the sparsity pattern matrix of $\mathbf{\Psi}^*$ or $\mathbf{\Theta}^*$. Our next theorem provides an improved rate of convergence using this newly defined notion of sparsity pattern.

**Theorem 3.8** (Improved Convergence Rate using Sparsity Pattern)**.** Suppose that $n \gtrsim (s + s_{\max}^2)\log d$ and take $\lambda$ such that $\lambda \asymp \sqrt{\log d/n}$. Let $T \gtrsim \log s$. Under Assumptions 3.1, 3.2 and 3.3, we have

$$\big\|\widehat{\mathbf{\Psi}}^{(T)} - \mathbf{\Psi}^*\big\|_2 = \mathcal{O}_{\mathbb{P}}\bigg(\|\mathbf{M}^*\|_2\sqrt{\frac{1}{n}}\bigg), \text{ and}$$
$$\big\|\widetilde{\mathbf{\Theta}}^{(T)} - \mathbf{\Theta}^*\big\|_2 = \mathcal{O}_{\mathbb{P}}\bigg(\|\mathbf{M}^*\|_2\sqrt{\frac{1}{n}} \vee \sqrt{\frac{\log d}{n}}\bigg).$$

*Proof of Theorem 3.8.* The proof is deferred to Appendix B in the supplementary material. □

Theorem 3.8 suggests that the rates of convergence can be bounded using the spectral norm of the sparsity pattern matrix $\mathbf{M}^*$, which are sometimes much sharper than those provided in Theorems 3.5 and 3.6. To demonstrate this observation, we consider a sequence of chain graphs specified by the following sparsity pattern matrices:

$$\mathbf{M}_k^c = \begin{bmatrix} \mathbf{A}_k & \mathbf{0} \\ \mathbf{0} & \mathbf{I}_{d-k-1} \end{bmatrix}, \quad \text{for} \quad k = 4, \ldots, 50,$$

where $\mathbf{A}_k \in \mathbb{R}^{(k+1)\times(k+1)}$ such that the $(i,j)$-th entry $A_{k,ij} = 1$ if $|i-j| \le 1$, and $A_{k,ij} = 0$ otherwise. $\mathbf{I}_{d-k-1} \in \mathbb{R}^{(d-k-1)\times(d-k-1)}$ is the identity matrix. Let $s_k$ be the total sparsity of $\mathbf{M}_k^c$, that is $s_k = 2k$. We plot the ratio of the two rates of convergence for estimating $\mathbf{\Psi}^*$ in Theorems 3.5 and 3.8, $\|\mathbf{M}_k^c\|_2^2/s_k$, versus $s_k$ in Figure 1. From Figure 1, we can see that the ratio goes to 0 as the total sparsity increases. This demonstrates that the convergence rate in Theorem 3.8 is indeed much sharper than that in Theorem 3.5, as least for the chain graphs constructed above. We also observe similar but less significant improvement for star-shape graphs. In Figure 2, we give an geometric illustration of the star and chain graphs.

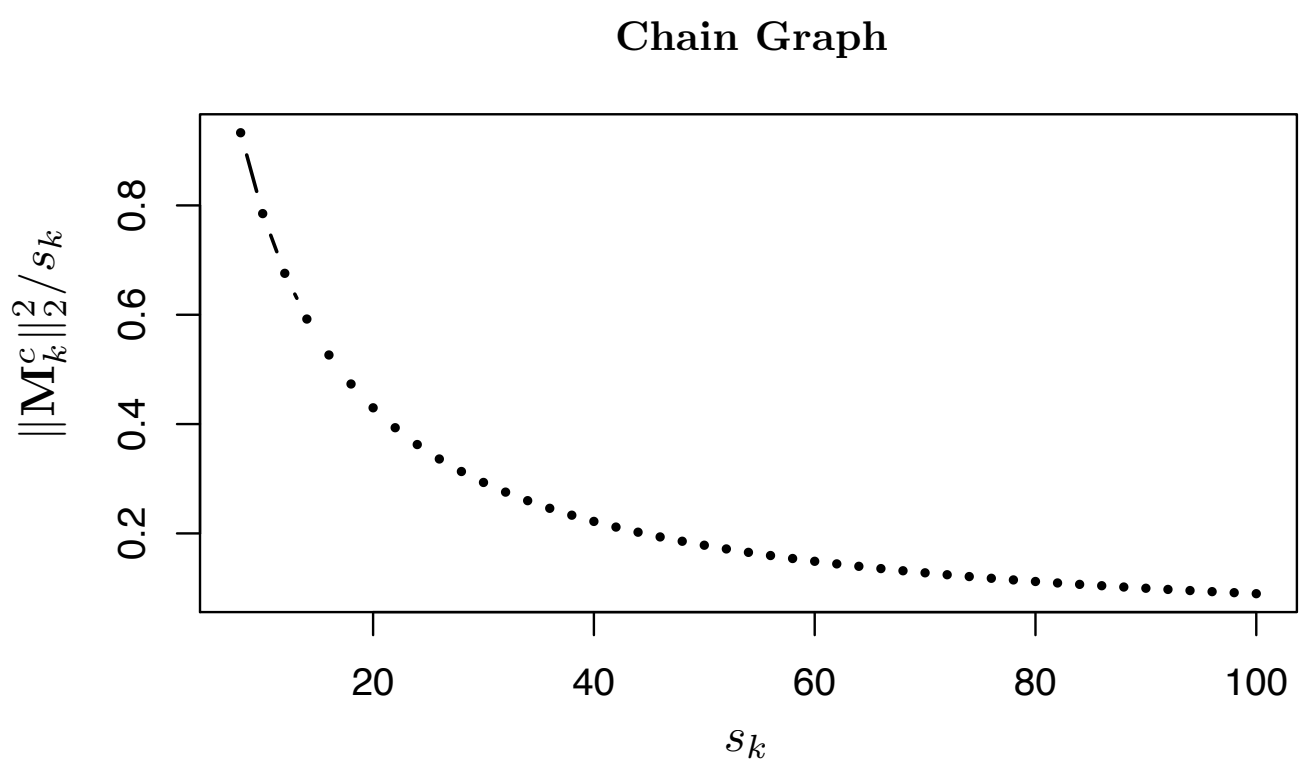

Figure 1: Convergence rates using sparsity pattern matrix $\mathbf{M}_k^c$ and total sparsity $s_k$.

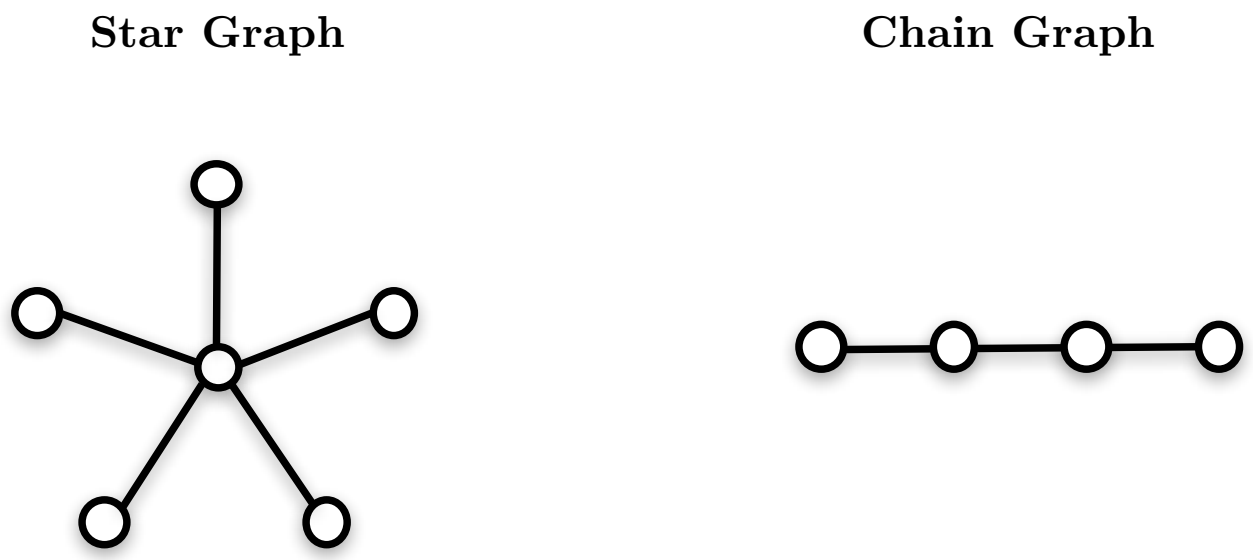

Figure 2: An illustration of the star and chain graphs.

# 4 Extension to Semiparametric Graphical Models

In this section, we extend the proposed method to modeling semiparametric graphical models. We focus on the nonparanormal family proposed by Liu et al. (2012), which is a nonparametric extension of the normal family. More specifically, we replace the random variable $\boldsymbol{X} = (X_1, \ldots, X_d)^{\mathrm{T}}$ by the transformation variable $f(\boldsymbol{X}) = (f_1(X_1), \ldots, f_d(X_d))^{\mathrm{T}}$, and assume that $f(\boldsymbol{X})$ follows a multivariate Gaussian distribution.

**Definition 4.1** (Nonparanormal)**.** Let $f = \{f_1, \ldots, f_d\}^{\mathrm{T}}$ be a set of monotone univariate functions and let $\boldsymbol{\Sigma}^{\mathrm{npn}} \in \mathbb{R}^{d\times d}$ be a positive-definite correlation matrix with $\mathrm{diag}(\boldsymbol{\Sigma}^{\mathrm{npn}}) = \mathbf{1}$. A $d$-dimensional random variable $\boldsymbol{X} = (X_1, \ldots, X_d)^{\mathrm{T}}$ has a nonparanormal distribution $\boldsymbol{X} \sim \mathrm{NPN}_d(f, \boldsymbol{\Sigma}^{\mathrm{npn}})$ if $f(\boldsymbol{X}) \equiv (f(X_1), \ldots, f_d(X_d))^{\mathrm{T}} \sim N_d(\mathbf{0}, \boldsymbol{\Sigma}^{\mathrm{npn}})$.

We aim to recover the precision matrix $\boldsymbol{\Theta}^{\mathrm{npn}} = (\boldsymbol{\Sigma}^{\mathrm{npn}})^{-1}$. The main idea behind this procedure is to exploit Kendall's tau statistics to directly estimate $\boldsymbol{\Theta}^{\mathrm{npn}}$, without explicitly calculating the marginal transformation functions $\{f_j\}_{j=1}^d$. We consider the following Kendall's tau statistic:

$$\widehat{\tau}_{jk} = \frac{2}{n(n-1)} \sum_{1\le i<i'\le n} \mathrm{sign}\big((X_j^{(i)} - X_j^{(i')})(X_k^{(i)} - X_k^{(i')})\big).$$

The Kendall's tau statistic $\widehat{\tau}_{jk}$ represent the nonparametric correlations between the empirical realizations of random variables $X_j$ and $X_k$ and is invariant to monotone trans-formations. Let $\widetilde{X}_j$ and $\widetilde{X}_k$ be two independent copies of $X_j$ and $X_k$. The population version of Kendall's tau is given by $\tau_{jk} \equiv \mathrm{Corr}\big(\mathrm{sign}(X_j - \widetilde{X}_j), \mathrm{sign}(X_k - \widetilde{X}_k)\big)$. We need the following lemma which is taken from Liu et al. (2012). It connects the Kendall's tau statistics to the underlying Pearson correlation coefficient $\mathbf{\Sigma}^{\mathrm{npn}}$.

**Lemma 4.2.** Assuming $\boldsymbol{X} \sim \mathrm{NPN}_d(f, \mathbf{\Sigma})$, we have $\Sigma^0_{jk} = \sin\big(\tau_{jk}\cdot\pi/2\big)$.

Motivated by this Lemma, we define the following estimators $\widehat{\boldsymbol{S}} = [\widehat{S}_{jk}]$ for the unknown correlation matrix $\mathbf{\Sigma}^{\mathrm{npn}}$:

$$\widehat{S}^{\tau}_{jk} = \begin{cases} \sin\big(\widehat{\tau}_{jk}\cdot\pi/2\big), & j \neq k, \\ 1, & j = k. \end{cases}$$

Now we are ready to prove the optimal spectral norm rate for the Gaussian copula graphical model. The results are provided in the following theorem.

**Theorem 4.3.** Assume that $n \gtrsim s\log d$ and let $\lambda \asymp \sqrt{\log d/n}$. Under Assumptions 3.1, 3.2 and 3.3, $\widehat{\mathbf{\Theta}}^{(\ell)}$ satisfies the following contraction property:

$$\big\|\widehat{\mathbf{\Theta}}^{(\ell)} - \mathbf{\Theta}^*\big\|_{\mathrm{F}} \le \underbrace{4\|\mathbf{\Theta}^*\|_2^2\|\nabla\mathcal{L}(\mathbf{\Theta}^*)_S\|_{\mathrm{F}}}_{\text{Optimal Rate}} + \underbrace{\frac{1}{2}\big\|\widehat{\mathbf{\Theta}}^{(\ell-1)} - \mathbf{\Theta}^*\big\|_{\mathrm{F}}}_{\text{Contraction}}, \quad 1 \le \ell \le T,$$

with probability at least $1 - 8/d$. If $T \gtrsim \log(\lambda\sqrt{n}) \gtrsim \log\log d$, we have

$$\big\|\widehat{\mathbf{\Theta}}^{(T)} - \mathbf{\Theta}^*\big\|_{\mathrm{F}} = \mathcal{O}_{\mathbb{P}}\left(\sqrt{\frac{s}{n}}\right).$$

*Proof of Theorem 4.3.* The proof is deferred to Appendix C in the supplementary material. □

## 5 Numerical Experiments

We compare our proposal to the graphical lasso (glasso) (Friedman et al., 2008) and neighborhood selection (NS) (Meinshausen and Bühlmann, 2006). Each of these approaches learns a Gaussian graphical model *via* an $\ell_1$ penalty on each edge. To evaluate the performance across different methods, we define the true positive rate as the proportion of correctly identified edges in the graph, and the false positive rate as the proportion of incorrectly identified edges in the graph. In addition, we calculate the difference between the estimated and true concentration matrix under the Frobenius norm. We do not compute this quantity for the NS approach since they do not estimate the concentration matrix directly.

For our proposal, we consider $T = 4$ iterations with the SCAD penalty proposed by Fan and Li (2001) that takes the following form:

$$p'_\lambda(t) = \begin{cases} \lambda & \text{if } |t| \le \lambda, \\ \frac{\gamma\lambda - |t|}{\gamma - 1} & \text{if } \lambda < |t| < \gamma\lambda, \\ 0 & \text{otherwise,} \end{cases}$$

where $\gamma > 2$. In all of our simulation studies, we pick $\gamma = 2.1$. Each of the methods involves a sparsity tuning parameter: we applied a fine grid of tuning parameter values to obtain the curves shown in Figure 3.

We consider cases with $n = \{150, 200\}$ and $d = 150$ with two set-ups for a $p \times p$ adjacency matrix $\mathbf{A}$: (i) random graph with 2.5% elements of $\mathbf{A}$ set to 1; (ii) band graph with $A_{i,i+1} = A_{i+1,i} = 1$ for $1 \leq i \leq d-1$. We then use the adjacency matrix $\mathbf{A}$ to create a matrix $\mathbf{E}$, as

$$E_{ij} = \begin{cases} 0 & \text{if } A_{ij} = 0 \\ 0.4 & \text{otherwise,} \end{cases}$$

and set $\mathbf{E} = \frac{1}{2}(\mathbf{E}+\mathbf{E}^T)$. Given the matrix $\mathbf{E}$, we set $\mathbf{\Theta}^{-1}$ equal to $\mathbf{E}+(0.1-e_{\min})\mathbf{I}$, where $e_{\min}$ is the smallest eigenvalue of $\mathbf{E}$. We then standardize the matrix $\mathbf{\Theta}^{-1}$ so that the diagonals are equal to one. Finally, we generate the data according to $\boldsymbol{X}^{(1)}, \ldots, \boldsymbol{X}^{(n)} \stackrel{\text{i.i.d.}}{\sim} N(\mathbf{0}, \mathbf{\Sigma})$. We present the results averaged over 100 data sets for each of the two simulation settings with $n = \{150, 200\}$ and $p = 150$ in Figure 3.

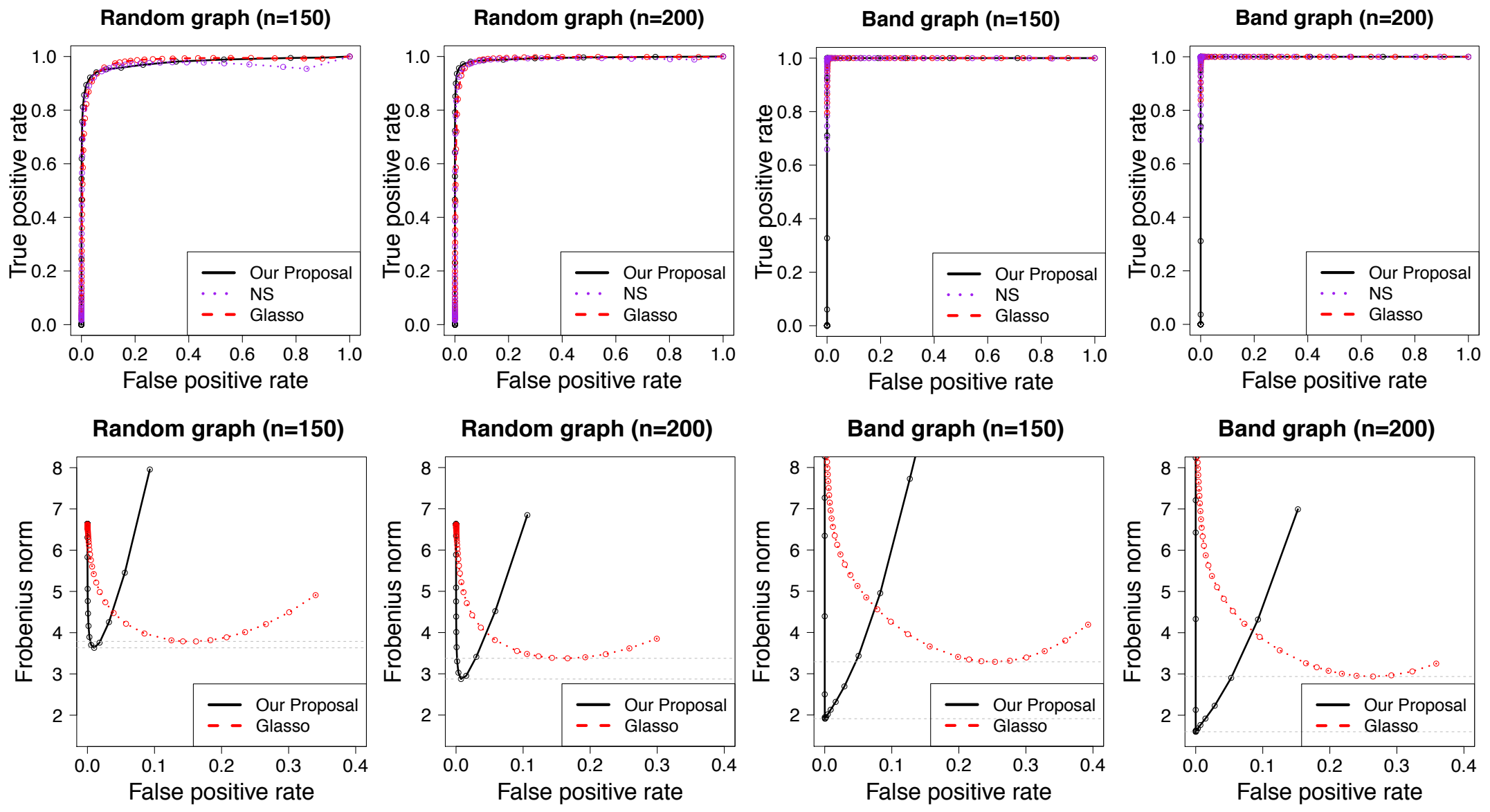

Figure 3: Row I: True and false positive rates, averaged over 100 data sets with $p = 150$, for random and band graphs, respectively. Row II: Difference between the estimated and the true inverse covariance matrices under the Frobenius norm. The different curves are obtained by varying the sparsity tuning parameter for each of the methods.

From Row I of Figure 3, we see that our proposal is very competitive relative to the existing proposals for estimating Gaussian graphical models in terms of true and false positive rates across all simulation settings. Row II of Figure 3 contains the difference between the estimated and the true inverse covariance matrices under the Frobenius norm as a function of the false positive rate. For random graph with $n = 150$, we see that the minimum error under the Frobenius norm for our proposal is smaller than that of the graphical lasso. As we increase the number of observations to $n = 200$, the difference between the minimum error for the two proposals are more apparent. More interestingly, the region for which our proposal has lower Frobenius norm than the graphical lasso is the primary region of interest. This is because an ideal estimator is one that has a low false positive rate while maintaining a high true positive rate with low error under the Frobenius norm. In contrast, the region for which the graphical lasso does better under the Frobenius norm is not the primary region of interest due to the high false positive rate. We see similar results for the band graph setting.

# 6 Conclusion and Discussions

We propose the graphical nonconvex optimization, which is then approximated by a sequence of convex programs, for estimating the inverse correlation and concentration matrices with better rates of convergence comparing with existing approaches. The proposed methodology is sequential convex in nature and thus is computationally tractable. Yet surprisingly, it produces estimators with oracle rate of convergence as if the global optimum for the penalized nonconvex problem could be obtained. Statistically, a contraction property is established: each convex program contracts the previous estimator by a 0.5-fraction until the optimal statistical error is reached.

Our work can be applied to many different topics: low rank matrix completion problems, high-dimensional quantile regression and many others. We conjecture that in all of the aforementioned topics, a similar sequential convex approximation can be proposed and can possibly give faster rate, with controlled computing resources. It is also interesting to see how our algorithm works in large-scale distributed systems. Is there any fundamental tradeoffs between statistical efficiency, communication and algorithmic complexity? We leave these as future research projects.

# Supplementary Material to "Graphical Nonconvex Optimization for Optimal Estimation in Gaussian Graphical Models"

Qiang Sun, Kean Ming Tan, Han Liu and Tong Zhang

**Abstract**

This supplementary material collects proofs for the main theoretical results in the main text and additional technical lemmas. The proofs of Proposition 3.4, Theorems 3.5 and 3.6 are collected in Section A. Section B provides the proof for Theorem 3.8. Proofs related to semiparametric graphical models are given in Section C. Various concentration inequalities and preliminary lemmas are postponed to Sections D and E, respectively.

## A Rate of Convergence in Frobenius Norm

This section presents an upper bound for the adaptive estimator $\widehat{\mathbf{\Psi}}^{(\ell)}$ in Frobenius norm, which in turn helps establish the scaling conditions needed to achieve the optimal spectral norm convergence rate.

### A.1 Proofs of Proposition 3.4, Theorems 3.5 and 3.6

In this section, we collect the proofs for Proposition 3.4, Theorems 3.5 and 3.6.

In order to suppress the noise at the $\ell$th step, it is necessary to control $\min_{(i,j)\in S}\big|\widehat{\Psi}_{ij}^{(\ell-1)}\big|$ in high dimensions. For this, we construct an entropy set, $\mathcal{E}_\ell$, of $S$ and analyze the magnitude of $\big\|\boldsymbol{\lambda}_{\mathcal{E}_\ell^c}^{(\ell-1)}\big\|_{\min}$. The entropy set at the $\ell$-th stage, $\mathcal{E}_\ell$, is defined as

$$\mathcal{E}_\ell=\Big\{(i,j):(i,j)\in S \text{ or } \lambda_{ij}^{(\ell-1)}<\lambda \mathrm{w}(u),\ \text{for } u=2\big(32\|\mathbf{\Psi}^*\|_2^2+\|\mathbf{\Sigma}^*\|_\infty^2\vee 1\big)\lambda\Big\}. \quad \text{(A.1)}$$

Thus the constant in Assumption 3.3 is $c=2(32\|\mathbf{\Psi}^*\|_2^2+\|\mathbf{\Sigma}^*\|_\infty^2\vee 1)$. Then it can be seen that $S\subseteq\mathcal{E}_\ell$, and thus $\mathcal{E}_\ell$ is an entropy set of $S$ for any $\ell\geq 1$. Proposition 3.4 follows from a slightly more general result below, which establishes rate of convergence for the one-step estimator of sparse inverse correlation matrix $\widehat{\mathbf{\Psi}}^{(1)}$.

**Proposition A.1** (One-step Estimator)**.** Assume that assumption 3.1 holds. Suppose $8\|\mathbf{\Psi}^*\|_2^2\lambda\sqrt{s}<1$. Take $\lambda$ such that $\lambda\asymp\sqrt{(\log d)/n}$ and suppose $n\gtrsim\log d$. Then with probability at least $1-8/d$, $\widehat{\mathbf{\Psi}}^{(1)}$ must satisfy

$$\big\|\widehat{\mathbf{\Psi}}^{(1)}-\mathbf{\Psi}^*\big\|_{\mathrm{F}}\leq C\|\mathbf{\Psi}^*\|_2^2\sqrt{\frac{s\log d}{n}}.$$

*Proof of Proposition A.1.* Define the event $\mathcal{J} = \{\|\widehat{\mathbf{C}} - \mathbf{C}^*\|_{\max} \le \lambda/2\}$. Then in the event $\mathcal{J}$, by applying Lemma A.4 and taking $\mathcal{E} = S$, we obtain $\|\widehat{\mathbf{\Psi}}^{(1)} - \mathbf{\Psi}^*\|_{\mathrm{F}} \le 4\|\mathbf{\Psi}^*\|_2^2 \cdot \lambda\sqrt{s}$. If we further take $\lambda = \sqrt{3c_2^{-1}}\sqrt{(\log d)/n} \asymp \sqrt{(\log d)/n}$, then by Lemma D.5, we have event $\mathcal{J}$ hold with probability at least $1 - 8d^{-1}$. The result follows by plugging the choice of $\lambda$. □

Theorems 3.5 and 3.6 follow form a slightly more general result below, which characterizes the rate of convergence of $\widehat{\mathbf{\Psi}}^{(\ell)}$ in Frobenius norm and that of $\widetilde{\mathbf{\Theta}}^{(T)}$ in spectral norm.

**Theorem A.2.** Assume that assumptions 3.1, 3.2 and 3.3. Suppose that $8\|\mathbf{\Psi}^*\|_2^2\lambda\sqrt{s} < 1$. Take $\lambda$ such that $\lambda \asymp \sqrt{\log d/n}$. Then with probability at least $1-8d^{-1}$, $\widehat{\mathbf{\Psi}}^{(\ell)}$ satisfies

$$\big\|\widehat{\mathbf{\Psi}}^{(\ell)} - \mathbf{\Psi}^*\big\|_{\mathrm{F}} \le \underbrace{8\|\mathbf{\Psi}^*\|_2^2\|\nabla\mathcal{L}(\mathbf{\Psi}^*)_S\|_{\mathrm{F}}}_{\text{Optimal Rate}} + \underbrace{\frac{1}{2}\big\|\widehat{\mathbf{\Psi}}^{(\ell-1)} - \mathbf{\Psi}^*\big\|_{\mathrm{F}}}_{\text{Contraction}}, \quad 1 \le \ell \le T.$$

Moreover, if that $T \gtrsim \log(\lambda\sqrt{n})$, we have $\big\|\widehat{\mathbf{\Psi}}^{(T)} - \mathbf{\Psi}^*\big\|_{\mathrm{F}} = \mathcal{O}_{\mathbb{P}}\big(\|\mathbf{\Psi}^*\|_2^2\sqrt{s/n}\big)$, and

$$\big\|\widetilde{\mathbf{\Theta}}^{(T)} - \mathbf{\Theta}^*\big\|_2 = \mathcal{O}_{\mathbb{P}}\bigg(\frac{\sigma_{\max}^3\|\mathbf{\Psi}^*\|_2}{\sigma_{\min}^3}\sqrt{\frac{\log d}{n}} \bigvee \frac{\|\mathbf{\Psi}^*\|_2^2}{\sigma_{\min}^2}\sqrt{\frac{s}{n}}\bigg).$$

*Proof of Theorem A.2.* Under the conditions of theorem, combining Proposition A.7 and Lemma D.5, we obtain the following contraction property of the solutions, $\{\widehat{\mathbf{\Psi}}^{(\ell)}\}_{\ell=1}^{\mathrm{T}}$,

$$\big\|\widehat{\mathbf{\Psi}}^{(\ell)} - \mathbf{\Psi}^*\big\|_{\mathrm{F}} \le 4\|\mathbf{\Psi}^*\|_2^2\|\nabla\mathcal{L}(\mathbf{\Psi}^*)_S\|_{\mathrm{F}} + \frac{1}{2}\big\|\widehat{\mathbf{\Psi}}^{(\ell-1)} - \mathbf{\Psi}^*\big\|_{\mathrm{F}}.$$

Next, we introduce an inequality by induction analysis. Specifically, if $a_n \le a_0 + \alpha a_{n-1}$, $\forall\, n \ge 2$ and $0 \le \alpha < 1$, then

$$a_n \le a_0\frac{1-\alpha^{n-1}}{1-\alpha} + \alpha^{n-1}a_1.$$

Taking $a_0 = 4\|\mathbf{\Psi}^*\|_2^2\|\nabla\mathcal{L}(\mathbf{\Psi}^*)_S\|_{\mathrm{F}}$, we obtain that $\big\|\widehat{\mathbf{\Psi}}^{(\ell)} - \mathbf{\Psi}^*\big\|_{\mathrm{F}} \le 8\|\mathbf{\Psi}^*\|_2^2\|\nabla\mathcal{L}(\mathbf{\Psi}^*)_S\|_{\mathrm{F}} + \big(1/2\big)^{\ell-1}\big\|\widehat{\mathbf{\Psi}}^{(1)} - \mathbf{\Psi}^*\big\|_{\mathrm{F}}$. In the sequel, we bound $\|\nabla\mathcal{L}(\mathbf{\Psi}^*)_S\|_{\mathrm{F}}$ and $\|\widehat{\mathbf{\Psi}}^{(1)} - \mathbf{\Psi}^*\|_{\mathrm{F}}$ respectively. By Proposition A.1, we have $\|\widehat{\mathbf{\Psi}}^{(1)} - \mathbf{\Psi}^*\|_{\mathrm{F}} \lesssim 8\|\mathbf{\Psi}^*\|_2^2\lambda\sqrt{s}$. Moreover, if we let $T \ge \log(\lambda\sqrt{n})/\log 2 \gtrsim \log(\lambda\sqrt{n})$, then $(1/2)^{\mathrm{T}-1}\|\widehat{\mathbf{\Psi}}^{(1)} - \mathbf{\Psi}^*\|_{\mathrm{F}} \le 16\|\mathbf{\Psi}^*\|_2^2 \cdot \sqrt{s/n}$. On the other side, we have $\|\nabla\mathcal{L}(\mathbf{\Psi}^*)_S\|_{\mathrm{F}} = \mathcal{O}_{\mathbb{P}}(\|\mathbf{\Psi}^*\|_2^2 \cdot \sqrt{s/n})$, which follows from Lemma D.4. Therefore, combining the above results obtains us that $\|\widehat{\mathbf{\Psi}}^{(T)} - \mathbf{\Psi}^*\|_{\mathrm{F}} = \mathcal{O}_{\mathbb{P}}\big(\|\mathbf{\Psi}^*\|_2^2\sqrt{s/n}\big)$.

To achieve the statistical rate for $\|\widetilde{\mathbf{\Theta}}^{(T)} - \mathbf{\Theta}^*\|_2$, we apply Lemma E.3 and obtain that

$$\begin{aligned}
\|\widetilde{\mathbf{\Theta}}^{(T)} - \mathbf{\Theta}^*\|_2 = &\big\|\big(\widehat{\mathbf{W}}^{-1} - \mathbf{W}^{-1}\big)\big(\widehat{\mathbf{\Psi}}^{(T)} - \mathbf{\Psi}^*\big)\big(\widehat{\mathbf{W}}^{-1} - \mathbf{W}^{-1}\big)\big\|_2 + \big\|\big(\widehat{\mathbf{W}}^{-1} - \mathbf{W}^{-1}\big)\widehat{\mathbf{\Psi}}^{(T)}\mathbf{W}^{-1}\big\|_2 \\
&+ \big\|\big(\widehat{\mathbf{W}}^{-1} - \mathbf{W}^{-1}\big)\mathbf{\Psi}^*\widehat{\mathbf{W}}^{-1}\big\|_2 + \big\|\widehat{\mathbf{W}}^{-1}\big(\widehat{\mathbf{\Psi}}^{(T)} - \mathbf{\Psi}^*\big)\mathbf{W}^{-1}\big\|_2 \\
\le &\underbrace{\|\widehat{\mathbf{W}}^{-1} - \mathbf{W}^{-1}\|_2^2\|\widehat{\mathbf{\Psi}}^{(T)} - \mathbf{\Psi}^*\|_2}_{\text{(R1)}} + \underbrace{\|\widehat{\mathbf{W}}^{-1} - \mathbf{W}^{-1}\|_2\|\widehat{\mathbf{\Psi}}^{(T)}\|_2\|\mathbf{W}^{-1}\|_2}_{\text{(R2)}} \\
&+ \underbrace{\|\widehat{\mathbf{W}}^{-1} - \mathbf{W}^{-1}\|_2\|\mathbf{\Psi}^*\|_2\|\widehat{\mathbf{W}}^{-1}\|_2}_{\text{(R3)}} + \underbrace{\|\widehat{\mathbf{W}}^{-1}\|_2\|\mathbf{W}^{-1}\|_2\|\widehat{\mathbf{\Psi}}^{(T)} - \mathbf{\Psi}^*\|_2}_{\text{(R4)}}.
\end{aligned}$$

We now bound terms (R1) to (R4) respectively. Before we proceed, we apply Lemma D.2 and the union sum bound to obtain that, for any $\varepsilon \geq 0$,

$$\mathbb{P}\Big(\|\widehat{\mathbf{W}}^2 - \mathbf{W}^2\|_2 > \varepsilon \max_i \Sigma^*_{ii}\Big) \leq d \cdot \exp\Big\{-n \cdot C(\varepsilon)\Big\} = \exp\Big\{-n \cdot C(\varepsilon) + \log d\Big\},$$

where $C(\varepsilon) = 2^{-1}(\varepsilon - \log(1+\varepsilon))$. Suppose that $0 \leq \varepsilon \leq 1/2$, then we have $-n \cdot C(\varepsilon) \leq -n \cdot \varepsilon^2/3$. Further suppose that $n \geq 36 \log d$ and take $\varepsilon = 3\sqrt{(\log d)/n}$, we obtain that $-n \cdot C(\varepsilon) + \log d \leq 2 \log d$ and

$$\mathbb{P}\left(\|\widehat{\mathbf{W}}^2 - \mathbf{W}^2\|_2 > 3\sigma^2_{\max} \cdot \sqrt{\frac{\log d}{n}}\right) \leq \frac{1}{d^2},$$

where we use the assumption that $\max_i \mathbf{\Sigma}^*_{ii} \leq \sigma^2_{\max}$. Therefore, we have $\big\|\widehat{\mathbf{W}}^2 - \mathbf{W}^2\big\|_2 = \mathcal{O}_{\mathbb{P}}\Big(\sigma^2_{\max} \cdot \sqrt{\log d/n}\Big)$. Since $\widehat{\mathbf{W}}^2$ and $\mathbf{W}^2$ are diagonal and thus commutative. We note that, for any two event $\mathcal{A}$ and $\mathcal{B}$, $\mathbb{P}(\mathcal{A}) = \mathbb{P}(\mathcal{A} \cap \mathcal{B}) + \mathbb{P}(\mathcal{A} \cap \mathcal{B}^c)$ holds. Therefore, for any $M > 0$, we have

$$\begin{aligned}
&\mathbb{P}\left(\big\|\widehat{\mathbf{W}}^{-1} - \mathbf{W}^{-1}\big\|_2 > M\sigma^2_{\max}\sqrt{\frac{\log d}{n}}\right) \\
&\quad \leq \mathbb{P}\Bigg(\big\|\widehat{\mathbf{W}}^{-1} - \mathbf{W}^{-1}\big\|_2 > M\sigma^2_{\max}\sqrt{\frac{\log d}{n}}, \\
&\qquad\qquad \big\|\widehat{\mathbf{W}}^{-1} - \mathbf{W}^{-1}\big\|_2 \leq 2(\sqrt{2}+1)\big\|\mathbf{W}\big\|_2 \lambda^{-2}_{\min}(\mathbf{W}^2)\big\|\widehat{\mathbf{W}}^2 - \mathbf{W}^2\big\|_2\Bigg) \\
&\qquad + \mathbb{P}\Big(\big\|\widehat{\mathbf{W}}^{-1} - \mathbf{W}^{-1}\big\|_2 > 2(\sqrt{2}+1)\big\|\mathbf{W}\big\|_2 \lambda^{-2}_{\min}(\mathbf{W}^2)\big\|\widehat{\mathbf{W}}^2 - \mathbf{W}^2\big\|_2\Big).
\end{aligned}$$

Further using Lemma E.7 yields that

$$\begin{aligned}
&\mathbb{P}\left(\big\|\widehat{\mathbf{W}}^{-1} - \mathbf{W}^{-1}\big\|_2 > M\sigma^2_{\max}\sqrt{\frac{\log d}{n}}\right) \\
&\quad \leq \underbrace{\mathbb{P}\left(2(\sqrt{2}+1)\big\|\mathbf{W}\big\|_2 \lambda^{-2}_{\min}(\mathbf{W}^2)\big\|\widehat{\mathbf{W}}^2 - \mathbf{W}^2\big\|_2 > M\sigma^2_{\max}\sqrt{\frac{\log d}{n}}\right)}_{\text{(T1)}} \\
&\qquad + \underbrace{\mathbb{P}\Big(\big\|\widehat{\mathbf{W}}^2 - \mathbf{W}^2\big\|_2 > 2^{-1}\lambda_{\min}(\mathbf{W}^2)\Big)}_{\text{(T2)}}.
\end{aligned}$$

By taking $M = M_1 \cdot \|\mathbf{W}\|_2 \lambda^{-2}_{\min}(\mathbf{W}^2) = M_1 \cdot \sigma_{\max}/\sigma^4_{\min}$ and letting $M_1 \to 0$, we get (T1) $\to 0$. Under the assumption that $\sigma^2_{\max}/\sigma^2_{\min} = O\big((n/\log d)^{1/3}\big)$, we have $\sigma^2_{\max}/\sigma^2_{\min} = o\big(\sqrt{n/\log d}\big)$, and thus (T2) $\to 0$. Therefore we obtain that $\big\|\widehat{\mathbf{W}}^{-1} - \mathbf{W}^{-1}\big\|_2 = \mathcal{O}_{\mathbb{P}}\big(\sigma^{-4}_{\min}\sigma^3_{\max}\sqrt{(\log d)/n}\big)$. Similarly, we have the following facts:

$$\big\|\widehat{\mathbf{\Psi}}^{(T)}\big\|_2 = \mathcal{O}_{\mathbb{P}}(\|\mathbf{\Psi}^*\|_2), \ \big\|\widehat{\mathbf{W}}^{-1}\big\|_2 = \lambda^{-1}_{\min}(\widehat{\mathbf{W}}) = \mathcal{O}_{\mathbb{P}}(\sigma^{-1}_{\min}), \text{ and } \big\|\mathbf{W}^{-1}\big\|_2 = \sigma^{-1}_{\min}.$$

Applying the above results to the terms (R1)-(R4). we obtain that

$$\text{(R1)} = \mathcal{O}_{\mathbb{P}}\left(\sigma^{-2}_{\min}\|\mathbf{\Psi}^*\|_2^2\sqrt{\frac{s}{n} \cdot \frac{\sigma^6_{\max}}{\sigma^6_{\min}}\frac{\log d}{n}}\right) = \mathcal{O}_{\mathbb{P}}\left(\sigma^{-2}_{\min}\|\mathbf{\Psi}^*\|_2^2\sqrt{\frac{s}{n}}\right),$$

$$\text{(R2)} = \text{(R3)} = \mathcal{O}_{\mathbb{P}}\left(\frac{\sigma^3_{\max}}{\sigma^3_{\min}}\|\mathbf{\Psi}^*\|_2\sqrt{\frac{\log d}{n}}\right), \ \text{(R4)} = \mathcal{O}_{\mathbb{P}}\left(\sigma^{-2}_{\min}\|\mathbf{\Psi}^*\|_2^2\sqrt{\frac{s}{n}}\right).$$

Therefore, by combining the rate for terms (R1)-(R4), we obtain the final result. $\square$

## A.2 Technical Lemmas

Define the symmetrized Bregman divergence for the loss function $\mathcal{L}(\cdot)$ as $D^s_{\mathcal{L}}(\boldsymbol{\Theta}, \boldsymbol{\Theta}^*) = \big\langle \nabla\mathcal{L}(\boldsymbol{\Theta}) - \mathcal{L}(\boldsymbol{\Theta}^*), \boldsymbol{\Theta} - \boldsymbol{\Theta}^* \big\rangle$. For any matrix $\mathbf{A} \in \mathbb{R}^{d\times d}$, let $\mathbf{A}_- \in \mathbb{R}^{d\times d}$ be the off diagonal matrix of $\mathbf{A}$ with diagonal entries equal to 0, and $\mathbf{A}_+ = \mathbf{A} - \mathbf{A}_-$ be the diagonal mtrix.

**Lemma A.3.** For the symmetrized Bregman divergence defined above, we have

$$D^s_{\mathcal{L}}(\boldsymbol{\Theta}, \boldsymbol{\Theta}^*) = \big\langle \nabla\mathcal{L}(\boldsymbol{\Theta}) - \nabla\mathcal{L}(\boldsymbol{\Theta}^*), \boldsymbol{\Theta} - \boldsymbol{\Theta}^* \big\rangle \geq \big(\|\boldsymbol{\Theta}^*\|_2 + \|\boldsymbol{\Theta} - \boldsymbol{\Theta}^*\|_2\big)^{-2} \|\boldsymbol{\Theta} - \boldsymbol{\Theta}^*\|_{\mathrm{F}}^2.$$

*Proof of Lemma A.3.* We use $\mathrm{vec}(\mathbf{A})$ to denote the vectorized form of any matrix $\mathbf{A}$. Then by the mean value theory, there exists a $\gamma \in [0, 1]$ such that,

$$\begin{aligned} D^s_{\mathcal{L}}(\boldsymbol{\Theta}, \boldsymbol{\Theta}^*) &= \big\langle \nabla\mathcal{L}(\boldsymbol{\Theta}) - \nabla\mathcal{L}(\boldsymbol{\Theta}^*), \boldsymbol{\Theta} - \boldsymbol{\Theta}^* \big\rangle = \mathrm{vec}(\boldsymbol{\Theta} - \boldsymbol{\Theta}^*)^{\mathrm{T}} \big(\nabla^2\mathcal{L}(\boldsymbol{\Theta}^* + \gamma\boldsymbol{\Delta})\big) \mathrm{vec}(\boldsymbol{\Theta} - \boldsymbol{\Theta}^*) \\ &\geq \lambda_{\min}\big(\nabla^2\mathcal{L}(\boldsymbol{\Theta}^* + \gamma\boldsymbol{\Delta})\big) \|\boldsymbol{\Delta}\|_{\mathrm{F}}^2, \quad \text{where } \boldsymbol{\Delta} = \boldsymbol{\Theta} - \boldsymbol{\Theta}^*. \end{aligned}$$

By standard properties of the Kronecker product and the Weyl's inequality (Horn and Johnson, 2012), we obtain that

$$\begin{aligned} \lambda_{\min}\Big(\nabla^2\mathcal{L}(\boldsymbol{\Theta}^* + \gamma\boldsymbol{\Delta})\Big) &= \lambda_{\min}\Big(\big((\boldsymbol{\Theta}^* + \gamma\boldsymbol{\Delta}) \otimes (\boldsymbol{\Theta}^* + \gamma\boldsymbol{\Delta})\big)^{-1}\Big) \\ &= \|\boldsymbol{\Theta}^* + \gamma\boldsymbol{\Delta}\|_2^{-2} \geq \big(\|\boldsymbol{\Theta}^*\|_2 + \gamma\|\boldsymbol{\Delta}\|_2\big)^{-2}. \end{aligned}$$

Finally, observing that $\gamma \leq 1$, we obtain

$$D^s_{\mathcal{L}}(\boldsymbol{\Theta}, \boldsymbol{\Theta}^*) = \big\langle \nabla\mathcal{L}(\boldsymbol{\Theta}) - \nabla\mathcal{L}(\boldsymbol{\Theta}^*), \boldsymbol{\Delta} \big\rangle \geq \big(\|\boldsymbol{\Theta}^*\|_2 + \|\boldsymbol{\Delta}\|_2\big)^{-2} \|\boldsymbol{\Theta} - \boldsymbol{\Theta}^*\|_{\mathrm{F}}^2.$$

Plugging the definition of $\boldsymbol{\Delta}$ obtains us the final bound. $\square$

The following lemma characterizes an upper bound of $\|\widehat{\boldsymbol{\Psi}} - \boldsymbol{\Psi}^*\|_{\mathrm{F}}$ by using localized analysis.

**Lemma A.4.** Suppose $8\|\boldsymbol{\Psi}^*\|_2\lambda\sqrt{s} < 1$. Take $\mathcal{E}$ such that $S \subseteq \mathcal{E}$ and $|\mathcal{E}| \leq 2s$. Further assume $\|\boldsymbol{\lambda}_{\mathcal{E}^c}\|_{\min} \geq \lambda/2 \geq \|\nabla\mathcal{L}(\boldsymbol{\Psi}^*)\|_{\max}$. Let $\widehat{\boldsymbol{\Psi}}$ be the solution to (B.4). Then $\widehat{\boldsymbol{\Psi}}$ must satisfy

$$\|\widehat{\boldsymbol{\Psi}} - \boldsymbol{\Psi}^*\|_{\mathrm{F}} \leq 4\|\boldsymbol{\Psi}^*\|_2^2\big(\|\boldsymbol{\lambda}_S\|_{\mathrm{F}} + \|\nabla\mathcal{L}(\boldsymbol{\Psi}^*)_{\mathcal{E}}\|_{\mathrm{F}}\big) \leq 8\|\boldsymbol{\Psi}^*\|_2^2\lambda\sqrt{s}.$$

*Proof of Lemma A.4.* We start by introducing an extra local parameter $r$ which satisfies $8\|\boldsymbol{\Psi}^*\|_2^2\lambda\sqrt{s} < r \leq \|\boldsymbol{\Psi}^*\|_2$. This is possible since $\lambda\sqrt{|\mathcal{E}|} \leq \sqrt{2}\lambda\sqrt{s} \to 0$ and $8\|\boldsymbol{\Psi}^*\|_2\lambda\sqrt{s} < 1$ by assumption. Based on this local parameter $r$, we construct an intermediate estimator: $\widetilde{\boldsymbol{\Psi}} = \boldsymbol{\Psi}^* + t\cdot(\widehat{\boldsymbol{\Psi}} - \boldsymbol{\Psi}^*)$, where $t$ is taken such that $\|(\widetilde{\boldsymbol{\Psi}} - \boldsymbol{\Psi}^*\|_{\mathrm{F}} = r$, if $\|(\widetilde{\boldsymbol{\Psi}} - \boldsymbol{\Psi}^*\|_{\mathrm{F}} > r$; $t = 1$ otherwise. Applying Lemma A.3 with $\boldsymbol{\Theta}_1 = \widetilde{\boldsymbol{\Psi}}$ and $\boldsymbol{\Theta}_2 = \boldsymbol{\Psi}^*$ obtains us

$$\big(\big\|\boldsymbol{\Psi}^*\big\|_2 + r\big)^{-2} \big\|\widetilde{\boldsymbol{\Psi}} - \boldsymbol{\Psi}^*\big\|_{\mathrm{F}}^2 \leq \big\langle \nabla\mathcal{L}(\widetilde{\boldsymbol{\Psi}}) - \nabla\mathcal{L}(\boldsymbol{\Psi}^*), \widetilde{\boldsymbol{\Psi}} - \boldsymbol{\Psi}^* \big\rangle. \tag{A.2}$$

To bound the right hand side of the above inequality, we use Lemma E.2 to obtain

$$D^s_{\mathcal{L}}(\widetilde{\boldsymbol{\Psi}}, \boldsymbol{\Psi}^*) \le tD^s_{\mathcal{L}}(\widehat{\boldsymbol{\Psi}}, \boldsymbol{\Psi}^*) = t\big\langle \nabla\mathcal{L}(\widehat{\boldsymbol{\Psi}}) - \nabla\mathcal{L}(\boldsymbol{\Psi}^*), \widehat{\boldsymbol{\Psi}} - \boldsymbol{\Psi}^*\big\rangle. \tag{A.3}$$

We note that the sub-differential of the norm $\|\cdot\|_{1,\mathrm{off}}$ evaluated at $\boldsymbol{\Psi}$ consists the set of all symmetric matrices $\boldsymbol{\Gamma} \in \mathbb{R}^{d\times d}$ such that $\Gamma_{ij} = 0$ if $i = j$; $\Gamma_{ij} = \mathrm{sign}(\Gamma_{ij})$ if $i \ne j$ and $\Psi_{ij} \ne 0$; $\Gamma_{ij} \in [-1, +1]$ if $i \ne j$ and $\Psi_{ij} = 0$, where $\Psi_{ij}$ is the $(i, j)$-th entry of $\boldsymbol{\Psi}$. Then by the Karush-Kuhn-Tucker conditions, there exists a $\widehat{\boldsymbol{\Gamma}} \in \partial\|\widehat{\boldsymbol{\Psi}}\|_{1,\mathrm{off}}$ such that $\nabla\mathcal{L}(\widehat{\boldsymbol{\Psi}})+\boldsymbol{\lambda}\odot\widehat{\boldsymbol{\Gamma}} = \widehat{\mathbf{C}} - \widehat{\boldsymbol{\Psi}}^{-1} + \boldsymbol{\lambda}\odot\widehat{\boldsymbol{\Gamma}} = \mathbf{0}$. Plugging (A.3) into (A.2) and adding the term $\langle\boldsymbol{\lambda}\odot\widehat{\boldsymbol{\Gamma}}, \widehat{\boldsymbol{\Psi}} - \boldsymbol{\Psi}^*\rangle$ on both sides of (A.3), we obtain

$$\begin{aligned}(\|\boldsymbol{\Psi}^*\|_2 + r)^{-2}\|\widetilde{\boldsymbol{\Psi}} - \boldsymbol{\Psi}^*\|_F^2 + t\underbrace{\langle\nabla\mathcal{L}(\boldsymbol{\Psi}^*), \widehat{\boldsymbol{\Psi}} - \boldsymbol{\Psi}^*\rangle}_{\mathrm{I}} + t\underbrace{\langle\boldsymbol{\lambda}\odot\widehat{\boldsymbol{\Gamma}}, \widehat{\boldsymbol{\Psi}} - \boldsymbol{\Psi}^*\rangle}_{\mathrm{II}} \\ \le t\underbrace{\langle\nabla\mathcal{L}(\widehat{\boldsymbol{\Psi}}) + \boldsymbol{\lambda}\odot\widehat{\boldsymbol{\Gamma}}, \widehat{\boldsymbol{\Psi}} - \boldsymbol{\Psi}^*\rangle}_{\mathrm{III}}.\end{aligned} \tag{A.4}$$

Next, we bound terms I, II and III respectively. For a set $\mathcal{E}$, let $\mathcal{E}^c$ denote its complement with respect to (w.r.t.) the full index set $\{(i, j) : 1 \le i, j \le d\}$. For term I, separating the support of $\nabla\mathcal{L}(\boldsymbol{\Psi})$ and $\widehat{\boldsymbol{\Psi}} - \boldsymbol{\Psi}^*$ to $\mathcal{E}\cup\mathcal{D}$ and $\mathcal{E}^c \setminus \mathcal{D}$, in which $\mathcal{D}$ is the set consisting of all diagonal elements, and then using the matrix Hölder inequality, we obtain

$$\begin{aligned}\big\langle\nabla\mathcal{L}(\boldsymbol{\Psi}^*), \widehat{\boldsymbol{\Psi}} - \boldsymbol{\Psi}^*\big\rangle &= \big\langle\big(\nabla\mathcal{L}(\boldsymbol{\Psi}^*)\big)_{\mathcal{E}\cup\mathcal{D}}, \big(\widehat{\boldsymbol{\Psi}} - \boldsymbol{\Psi}^*\big)_{\mathcal{E}\cup\mathcal{D}}\big\rangle + \big\langle\big(\nabla\mathcal{L}(\boldsymbol{\Psi}^*)\big)_{\mathcal{E}^c\setminus\mathcal{D}}, \big(\widehat{\boldsymbol{\Psi}} - \boldsymbol{\Psi}^*\big)_{\mathcal{E}^c\setminus\mathcal{D}}\big\rangle \\ &\ge -\big\|\big(\nabla\mathcal{L}(\boldsymbol{\Psi}^*)\big)_{\mathcal{E}\cup\mathcal{D}}\big\|_{\mathrm{F}}\big\|\big(\widehat{\boldsymbol{\Psi}} - \boldsymbol{\Psi}^*\big)_{\mathcal{E}\cup\mathcal{D}}\big\|_{\mathrm{F}} \\ &\quad - \big\|\big(\nabla\mathcal{L}(\boldsymbol{\Psi}^*)\big)_{\mathcal{E}^c\setminus\mathcal{D}}\big\|_{\mathrm{F}}\big\|\big(\widehat{\boldsymbol{\Psi}} - \boldsymbol{\Psi}^*\big)_{\mathcal{E}^c\setminus\mathcal{D}}\big\|_{\mathrm{F}}.\end{aligned}$$

For term II, separating the support of $(\boldsymbol{\lambda}\odot\widehat{\boldsymbol{\Gamma}})$ and $(\widehat{\boldsymbol{\Psi}} - \boldsymbol{\Psi}^*)$ to $S\cup\mathcal{D}$ and $S^c \setminus \mathcal{D}$, we obtain

$$\langle(\boldsymbol{\lambda}\odot\widehat{\boldsymbol{\Gamma}}), (\widehat{\boldsymbol{\Psi}} - \boldsymbol{\Psi}^*)\rangle = \langle(\boldsymbol{\lambda}\odot\widehat{\boldsymbol{\Gamma}})_{S\cup\mathcal{D}}, (\widehat{\boldsymbol{\Psi}} - \boldsymbol{\Psi}^*)_{S\cup\mathcal{D}}\rangle + \langle(\boldsymbol{\lambda}\odot\widehat{\boldsymbol{\Gamma}})_{S^c\setminus\mathcal{D}}, (\widehat{\boldsymbol{\Psi}} - \boldsymbol{\Psi}^*)_{S^c\setminus\mathcal{D}}\rangle. \tag{A.5}$$

For the last term in the above equality, we have

$$\langle(\boldsymbol{\lambda}\odot\widehat{\boldsymbol{\Gamma}})_{S^c\setminus\mathcal{D}}, (\widehat{\boldsymbol{\Psi}} - \boldsymbol{\Psi}^*)_{S^c\setminus\mathcal{D}}\rangle = \langle\boldsymbol{\lambda}_{S^c\setminus\mathcal{D}}, |\widehat{\boldsymbol{\Psi}}_{S^c\setminus\mathcal{D}}|\rangle = \langle\boldsymbol{\lambda}_{S^c\setminus\mathcal{D}}, |(\widehat{\boldsymbol{\Psi}} - \boldsymbol{\Psi}^*)_{S^c\setminus\mathcal{D}}|\rangle. \tag{A.6}$$

Plugging (A.6) into (A.5) and applying matrix Hölder inequality yields

$$\begin{aligned}\langle(\boldsymbol{\lambda}\odot\widehat{\boldsymbol{\Gamma}}, \widehat{\boldsymbol{\Psi}} - \boldsymbol{\Psi}^*\rangle &= \langle(\boldsymbol{\lambda}\odot\widehat{\boldsymbol{\Gamma}})_{S\cup\mathcal{D}}, (\widehat{\boldsymbol{\Psi}} - \boldsymbol{\Psi}^*)_{S\cup\mathcal{D}}\rangle + \langle\boldsymbol{\lambda}_{S^c\setminus\mathcal{D}}, |(\widehat{\boldsymbol{\Psi}} - \boldsymbol{\Psi}^*)_{S^c\setminus\mathcal{D}}|\rangle \\ &= \langle(\boldsymbol{\lambda}\odot\widehat{\boldsymbol{\Gamma}})_S, (\widehat{\boldsymbol{\Psi}} - \boldsymbol{\Psi})_S\rangle + \|\boldsymbol{\lambda}_{S^c\setminus\mathcal{D}}\|_{\mathrm{F}}\|(\widehat{\boldsymbol{\Psi}} - \boldsymbol{\Psi}^*)_{S^c\setminus\mathcal{D}}\|_{\mathrm{F}} \\ &\ge -\|\boldsymbol{\lambda}_S\|_{\mathrm{F}}\|(\widehat{\boldsymbol{\Psi}} - \boldsymbol{\Psi}^*)_S\|_{\mathrm{F}} + \|\boldsymbol{\lambda}_{\mathcal{E}^c\setminus\mathcal{D}}\|_{\mathrm{F}}\|(\widehat{\boldsymbol{\Psi}} - \boldsymbol{\Psi}^*)_{\mathcal{E}^c\setminus\mathcal{D}}\|_{\mathrm{F}},\end{aligned}$$

where we use $\boldsymbol{\lambda}_{\mathcal{D}} = \mathbf{0}$ in the second equality and $\mathcal{E}^c\setminus\mathcal{D} \subseteq S^c\setminus\mathcal{D}$ in the last inequality. For term III, using the optimality condition, we have $\mathrm{III} = \big\langle\nabla\mathcal{L}(\widehat{\boldsymbol{\Psi}}) + \boldsymbol{\lambda}\odot\widehat{\boldsymbol{\Gamma}}, \widehat{\boldsymbol{\Psi}} - \boldsymbol{\Psi}\big\rangle = 0$. Plugging the bounds for term I, II and III back into (A.4), we find that

$$\begin{aligned}\big(\|\boldsymbol{\Psi}^*\|_2 + r\big)^{-2}\big\|\widetilde{\boldsymbol{\Psi}} - \boldsymbol{\Psi}^*\big\|_{\mathrm{F}}^2 + t\big(\|\boldsymbol{\lambda}_{\mathcal{E}^c\setminus\mathcal{D}}\|_{\mathrm{F}} - \|(\nabla\mathcal{L}(\boldsymbol{\Psi}^*))_{\mathcal{E}^c\setminus\mathcal{D}}\|_{\mathrm{F}}\big)\cdot\big\|(\widehat{\boldsymbol{\Psi}} - \boldsymbol{\Psi}^*)_{\mathcal{E}^c\setminus\mathcal{D}}\big\|_{\mathrm{F}} \\ \le t\big(\big\|\big(\nabla\mathcal{L}(\boldsymbol{\Psi}^*)\big)_{\mathcal{E}\cup\mathcal{D}}\big\|_{\mathrm{F}} + \big\|\boldsymbol{\lambda}_S\big\|_{\mathrm{F}}\big)\cdot\big\|\widehat{\boldsymbol{\Psi}} - \boldsymbol{\Psi}^*\big\|_{\mathrm{F}}.\end{aligned}$$

Further observing the facts that $\|\boldsymbol{\lambda}_{\mathcal{E}^c\setminus\mathcal{D}}\|_F \geq \sqrt{|\mathcal{E}^c\setminus\mathcal{D}|}\big\|\boldsymbol{\lambda}_{\mathcal{E}^c\setminus\mathcal{D}}\big\|_{\min} \geq \sqrt{|\mathcal{E}^c\setminus\mathcal{D}|}\big\|\nabla\mathcal{L}(\boldsymbol{\Psi}^*)\big\|_{\max} \geq \big\|\big(\nabla\mathcal{L}(\boldsymbol{\Psi}^*)\big)_{\mathcal{E}^c\setminus\mathcal{D}}\big\|_{\mathrm{F}}$ and $t\|\widehat{\boldsymbol{\Psi}} - \boldsymbol{\Psi}^*\|_{\mathrm{F}} = \|\widetilde{\boldsymbol{\Psi}} - \boldsymbol{\Psi}^*\|_{\mathrm{F}}$, dividing both sides by $\|\widetilde{\boldsymbol{\Psi}} - \boldsymbol{\Psi}^*\|_{\mathrm{F}}$, we can simplify the above inequality to

$$(\|\boldsymbol{\Psi}^*\|_2+r)^{-2}\|\widetilde{\boldsymbol{\Psi}}-\boldsymbol{\Psi}^*\|_{\mathrm{F}} \leq \|\boldsymbol{\lambda}_S\|_{\mathrm{F}}+\|\nabla\mathcal{L}(\boldsymbol{\Psi}^*)_{\mathcal{E}\cup\mathcal{D}}\|_{\mathrm{F}} = \|\boldsymbol{\lambda}_S\|_{\mathrm{F}}+\|\nabla\mathcal{L}(\boldsymbol{\Psi}^*)_{\mathcal{E}}\|_{\mathrm{F}} \leq 2\lambda\sqrt{s},$$

where we use $\|\nabla\mathcal{L}(\boldsymbol{\Psi}^*)_{\mathcal{E}\cup\mathcal{D}}\|_{\mathrm{F}} = \|(\widehat{\mathbf{C}} - \mathbf{C}^*)_{\mathcal{E}\cup\mathcal{D}}\|_{\mathrm{F}} = \|(\widehat{\mathbf{C}} - \mathbf{C}^*)_{\mathcal{E}}\|_{\mathrm{F}} = \|\nabla\mathcal{L}(\boldsymbol{\Psi}^*)_{\mathcal{E}}\|_{\mathrm{F}}$ in the equality, and the last inequality follows from the Cauchy-Schwarz inequality, the fact $\|\boldsymbol{\lambda}\|_{\max} \leq \lambda$ and the assumption that $\lambda \geq 2\|\nabla\mathcal{L}(\boldsymbol{\Psi}^*)\|_{\max}$. Therefore, by the definition of $r$, we obtain $\|\widetilde{\boldsymbol{\Psi}} - \boldsymbol{\Psi}^*\|_{\mathrm{F}} \leq 2(\|\boldsymbol{\Psi}^*\|_2 + r)^2\lambda\sqrt{s} \leq 8\|\boldsymbol{\Psi}^*\|_2^2\lambda\sqrt{s} < r$, which implies $\widetilde{\boldsymbol{\Psi}} = \widehat{\boldsymbol{\Psi}}$ from the construction of $\widetilde{\boldsymbol{\Psi}}$. Thus $\widehat{\boldsymbol{\Psi}}$ satisfies the desired $\ell_2$ error bound. □

Recall the definition of $\mathcal{E}_\ell$, $1 \leq \ell \leq T$. We can bound $\|\widehat{\boldsymbol{\Psi}}^{(\ell)} - \boldsymbol{\Psi}^*\|_{\mathrm{F}}$ in terms of $\|\boldsymbol{\lambda}_S^{(\ell-1)}\|_{\mathrm{F}}$.

**Lemma A.5** (Sequential Bound). Under the same assumptions and conditions in Lemma A.4, for $\ell \geq 1$, $\widehat{\boldsymbol{\Psi}}^{(\ell)}$ must satisfy

$$\|\widehat{\boldsymbol{\Psi}}^{(\ell)} - \boldsymbol{\Psi}^*\|_{\mathrm{F}} \leq 4\big\|\boldsymbol{\Psi}^*\big\|_2^2\big(\big\|\boldsymbol{\lambda}_S^{(\ell-1)}\big\|_{\mathrm{F}} + \big\|\nabla\mathcal{L}(\boldsymbol{\Psi}^*)_{\mathcal{E}_\ell}\big\|_{\mathrm{F}}\big).$$

*Proof of Lemma A.5.* Now if we assume that for all $\ell \geq 1$, we have the following

$$|\mathcal{E}_\ell| \leq 2s, \quad \text{where } \ \mathcal{E}_\ell \ \text{ is defined in (A.1) , and} \tag{A.7}$$
$$\big\|\boldsymbol{\lambda}^{(\ell-1)}_{\mathcal{E}_\ell^c\setminus D}\big\|_{\min} \geq \lambda/2 \geq \|\nabla\mathcal{L}(\boldsymbol{\Psi}^*)\|_{\max}. \tag{A.8}$$

Using the matrix Hölder inequality, we obtain

$$\big\|\boldsymbol{\lambda}_S^{(\ell-1)}\big\|_{\mathrm{F}} \leq \sqrt{|S|}\big\|\boldsymbol{\lambda}_S\big\|_{\max} \leq \lambda\sqrt{s} \text{ and } \|\nabla\mathcal{L}(\boldsymbol{\Psi}^*)_{\mathcal{E}_\ell}\|_{\mathrm{F}} \leq \sqrt{|\mathcal{E}_\ell|}\|\nabla\mathcal{L}(\boldsymbol{\Psi}^*)_{\mathcal{E}_\ell}\|_{\max}.$$

Therefore, we have

$$\big\|\boldsymbol{\lambda}_S^{(\ell-1)}\big\|_{\mathrm{F}}+\big\|\nabla\mathcal{L}(\boldsymbol{\Psi}^*)_{\mathcal{E}_\ell}\big\|_{\mathrm{F}} \leq \lambda\sqrt{s}+\|\nabla\mathcal{L}(\boldsymbol{\Psi}^*)_{\mathcal{E}_\ell}\|_{\max}\sqrt{|\mathcal{E}_\ell|} \leq 2\lambda\sqrt{s}, \tag{A.9}$$

where the second inequality is due to the assumption that $\|\nabla\mathcal{L}(\boldsymbol{\Psi}^*)\|_{\max} \leq \lambda/2$. The $\ell_2$ error bound is given by Lemma A.4 by taking $\boldsymbol{\lambda} = \boldsymbol{\lambda}^{(\ell-1)}$ and $\mathcal{E} = \mathcal{E}_\ell$, i.e.

$$\big\|\widehat{\boldsymbol{\Psi}}^{(\ell)} - \boldsymbol{\Psi}^*\big\|_{\mathrm{F}} \leq 4\big\|\boldsymbol{\Psi}^*\big\|_2^2 \cdot \big(\big\|\boldsymbol{\lambda}_S^{(\ell-1)}\big\|_{\mathrm{F}} + \big\|\nabla\mathcal{L}(\boldsymbol{\Psi}^*)_{\mathcal{E}_\ell}\big\|_{\mathrm{F}}\big) \leq 8\|\boldsymbol{\Psi}^*\|_2^2 \cdot \lambda\sqrt{s}, \tag{A.10}$$

where last inequality is due to (A.9). Therefore, we only need to prove that (A.7) and (A.8) hold by induction. For $\ell = 1$, we have $\lambda \geq \lambda \mathrm{w}(u)$ for any $u$ and thus $\mathcal{E}_1 = S$, which implies that (A.7) and (A.8) hold for $\ell = 1$. Now assume that (A.7) and (A.8) hold at $\ell - 1$ for some $\ell \geq 2$. Since $(i,j) \in \mathcal{E}_\ell\setminus S$ implies that $(i,j) \notin S$ and $\lambda \mathrm{w}\big(\widehat{\Psi}_{ij}^{(\ell-1)}\big) = \lambda_j^{(\ell)} < \lambda\mathrm{w}(u) = \lambda/2$. By assumption, and since $\mathrm{w}(x)$ is non-increasing, we must have $\big|\widehat{\Psi}_{ij}^{(\ell-1)}\big| \geq u$. Therefore by induction hypothesis, we obtain that

$$\sqrt{|\mathcal{E}_\ell\setminus S|} \leq \frac{\big\|\widehat{\boldsymbol{\Psi}}^{(\ell-1)}_{\mathcal{E}_\ell\setminus S}\big\|_{\mathrm{F}}}{u} \leq \frac{\big\|\widehat{\boldsymbol{\Psi}}^{(\ell-1)} - \boldsymbol{\Psi}^*\big\|_{\mathrm{F}}}{u} \leq \frac{8\|\boldsymbol{\Psi}^*\|_2^2\lambda}{u} \cdot \sqrt{s} \leq \sqrt{s},$$

where the second last inequality follows from Lemma A.4, the fact that (A.7) and (A.8) hold at $\ell - 1$. This implies that $|\mathcal{E}_\ell| \leq 2|S| = 2s$. Now for such $\mathcal{E}_\ell^c$, we have $\|\boldsymbol{\lambda}_{\mathcal{E}_\ell^c}\|_{\min} \geq \lambda\mathrm{w}(u) \geq \lambda/2 \geq \|\nabla\mathcal{L}(\boldsymbol{\Psi})\|_\infty$, which completes the induction step. □

Our next lemma establishes the relationship between the adaptive regularization parameter $\boldsymbol{\lambda}$ and the estimator from the previous step.

**Lemma A.6.** Assume $\mathrm{w}(\cdot) \in \mathcal{T}$. Let $\lambda_{ij} = \lambda \mathrm{w}\big(|\Theta_{ij}|\big)$ for some $\boldsymbol{\Theta} = (\Theta_{ij})$ and $\mathrm{w}(\boldsymbol{\Theta}_S) = \big(\mathrm{w}(\Theta_{ij})\big)_{(i,j)\in S}$, then for the Frobenius norm $\|\cdot\|_{\mathrm{F}}$, we have

$$\big\|\boldsymbol{\lambda}_S\big\|_{\mathrm{F}} \le \lambda \big\|\mathrm{w}\big(|\boldsymbol{\Theta}_S^*| - u\big)\big\|_{\mathrm{F}} + \lambda u^{-1}\big\|\boldsymbol{\Theta}_S^* - \boldsymbol{\Theta}_S\big\|_{\mathrm{F}}.$$

*Proof of Lemma A.6.* By assumption, if $|\Theta_{ij}^* - \Theta_{ij}| \ge u$, then $\mathrm{w}\big(|\Theta_{ij}|\big) \le 1 \le u^{-1}|\Theta_{ij} - \Theta_{ij}^*|$; otherwise, $\mathrm{w}\big(|\Theta_{ij}|\big) \le \mathrm{w}\big(|\Theta_{ij}^*| - u\big)$. Therefore,the following inequality always hold:

$$\mathrm{w}\big(|\Theta_{ij}|\big) \le \mathrm{w}\big(|\Theta_{ij}^*| - u\big) + u^{-1}|\Theta_{ij}^* - \Theta_{ij}|.$$

Then by applying the $\|\cdot\|_*$-norm triangle inequality, we obtain that

$$\big\|\boldsymbol{\lambda}_S\big\|_{\mathrm{F}} \le \lambda \big\|\mathrm{w}\big(|\boldsymbol{\Theta}_S^*| - u\big)\big\|_{\mathrm{F}} + \lambda u^{-1}\big\|\boldsymbol{\Theta}_S^* - \boldsymbol{\Theta}_S\big\|_{?F}.$$

□

Our last technical result concerns a contraction property, namely, how the sequential approach improves the rate of convergence adaptively.

**Proposition A.7** (Contraction Property)**.** Assume that assumptions 3.1, 3.2 and 3.3 hold. Assume that $\lambda \ge 2\|\nabla \mathcal{L}(\boldsymbol{\Psi}^*)\|_{\max}$ and $8\|\boldsymbol{\Psi}^*\|_2^2 \lambda \sqrt{s} < 1$. Then $\widehat{\boldsymbol{\Psi}}^{(\ell)}$ satisfies the following contraction property

$$\big\|\widehat{\boldsymbol{\Psi}}^{(\ell)} - \boldsymbol{\Psi}^*\big\|_{\mathrm{F}} \le 4\|\boldsymbol{\Psi}^*\|_2^2 \|\nabla \mathcal{L}(\boldsymbol{\Psi}^*)_S\|_{\mathrm{F}} + \frac{1}{2}\big\|\widehat{\boldsymbol{\Psi}}^{(\ell-1)} - \boldsymbol{\Psi}^*\big\|_{\mathrm{F}}.$$

*Proof of Proposition A.7.* Under the conditions of the theorem, the proof of Lemma A.5 yields that

$$|\mathcal{E}_\ell| \le 2s, \quad \text{where } \ \mathcal{E}_\ell \text{ is defined in (A.1), and } \|\boldsymbol{\lambda}^{(\ell-1)}_{\mathcal{E}_\ell^c \setminus D}\|_{\min} \ge \|\nabla \mathcal{L}(\boldsymbol{\Psi}^*)\|_{\max}.$$

Thus, applying Lemma A.5 with $\widehat{\boldsymbol{\Psi}} = \widehat{\boldsymbol{\Psi}}^{(\ell)}, \boldsymbol{\lambda} = \boldsymbol{\lambda}^{(\ell-1)}$ and $\mathcal{E} = \mathcal{E}_\ell$, we obtain

$$\big\|\widehat{\boldsymbol{\Psi}}^{(\ell)} - \boldsymbol{\Psi}^*\big\|_{\mathrm{F}} \le 4\big\|\boldsymbol{\Psi}^*\big\|_2^2 \cdot \big(\|\boldsymbol{\lambda}_S^{(\ell-1)}\|_{\mathrm{F}} + \big\|\nabla \mathcal{L}(\boldsymbol{\Psi}^*)_{\mathcal{E}_\ell}\big\|_{\mathrm{F}}\big). \tag{A.11}$$

On the other side, by Lemma A.6, we can bound $\|\boldsymbol{\lambda}_S^{(\ell-1)}\|$ in terms of $\|\widehat{\boldsymbol{\Psi}}^{(\ell-1)} - \boldsymbol{\Psi}^*\|_{\mathrm{F}}$:

$$\big\|\boldsymbol{\lambda}_S^{(\ell-1)}\big\|_{\mathrm{F}} \le \lambda \big\|\mathrm{w}(|\boldsymbol{\Psi}_S^*| - u)\big\|_{\mathrm{F}} + \lambda u^{-1}\big\|\widehat{\boldsymbol{\Psi}}^{(\ell-1)} - \boldsymbol{\Psi}^*\big\|_{\mathrm{F}}. \tag{A.12}$$

Plugging the bound (A.12) into (A.11) yields that

$$\begin{aligned} \big\|\widehat{\boldsymbol{\Psi}}^{(\ell)} - \boldsymbol{\Psi}^*\big\|_{\mathrm{F}} \le 4\big\|\boldsymbol{\Psi}^*\big\|_2^2 \Big(\underbrace{\big\|\nabla \mathcal{L}(\boldsymbol{\Psi}^*)_{\mathcal{E}_\ell}\big\|_{\mathrm{F}}}_{\mathrm{I}} + \lambda \big\|\mathrm{w}(|\boldsymbol{\Psi}_S^*| - u)\big\|_{\mathrm{F}}\Big) \\ + 4\|\boldsymbol{\Psi}^*\|_2^2 \lambda u^{-1}\big\|\widehat{\boldsymbol{\Psi}}^{(\ell-1)} - \boldsymbol{\Psi}^*\big\|_{\mathrm{F}}. \end{aligned} \tag{A.13}$$

In the next, we bound term I. Separating the support of $\big(\nabla \mathcal{L}(\boldsymbol{\Psi}^*)\big)_{\mathcal{E}_\ell}$ to $S$ and $\mathcal{E}_\ell \setminus S$ and then using triangle inequality, we obtain

$$\mathrm{I} = \big\|\nabla \mathcal{L}(\boldsymbol{\Psi}^*)_{\mathcal{E}_\ell}\big\|_{\mathrm{F}} \le \big\|\nabla \mathcal{L}(\boldsymbol{\Psi}^*)_S\big\|_{\mathrm{F}} + \big\|\nabla \mathcal{L}(\boldsymbol{\Psi}^*)_{\mathcal{E}_\ell \setminus S}\big\|_{\mathrm{F}}. \tag{A.14}$$

Moreover, we have the following facts. First, we have $\|\nabla\mathcal{L}(\mathbf{\Psi}^*)_{\mathcal{E}_\ell\backslash S}\|_2 \le \sqrt{|\mathcal{E}_\ell\backslash S|}\|\nabla\mathcal{L}(\mathbf{\Psi}^*)\|_{\max}$ by the Hölder inequality. From the assumption, we know $\|\nabla\mathcal{L}(\mathbf{\Psi}^*)\|_{\max} \le \lambda/2$. Plugging these bounds into (A.14) results that $\|\nabla\mathcal{L}(\mathbf{\Psi}^*)_{\mathcal{E}_\ell}\|_{\mathrm{F}} \le \|\nabla\mathcal{L}(\mathbf{\Psi}^*)_S\|_{\mathrm{F}} + \lambda\sqrt{|\mathcal{E}_\ell\backslash S|}$. Now, by following a similar argument in Lemma A.5, we can bound $\sqrt{|\mathcal{E}_\ell\backslash S|}$ by $\|\widehat{\mathbf{\Psi}}^{(\ell-1)}_{\mathcal{E}_\ell\backslash S}\|_{\mathrm{F}}/u \le \|\widehat{\mathbf{\Psi}}^{(\ell-1)} - \mathbf{\Psi}^*\|_{\mathrm{F}}/u$. Therefore, term I can be bounded by $\|\nabla\mathcal{L}(\mathbf{\Psi}^*)_S\|_{\mathrm{F}} + \lambda u^{-1}\|\widehat{\mathbf{\Psi}}^{(\ell-1)} - \mathbf{\Psi}^*\|_{\mathrm{F}}$. Plugging the upper bound for I into (A.13), we obtain

$$\|\widehat{\mathbf{\Psi}}^{(\ell)} - \mathbf{\Psi}^*\|_{\mathrm{F}} \le 4\|\mathbf{\Psi}^*\|_2^2\Big(\|\nabla\mathcal{L}(\mathbf{\Psi}^*)_S\|_2 + \lambda\|\mathrm{w}(|\mathbf{\Psi}^*_S| - u)\|_2\Big) + (4\|\mathbf{\Psi}^*\|_2^2 + 1)\lambda u^{-1}\|\widehat{\mathbf{\Psi}}^{(\ell-1)} - \mathbf{\Psi}^*\|_{\mathrm{F}}.$$

Now observing that $\|\mathbf{\Psi}^*_S\|_{\min} \ge u + \alpha\lambda \asymp \lambda$, thus $\mathrm{w}(|\mathbf{\Theta}^*_S| - u) \le \mathrm{w}(\alpha\lambda\cdot\mathbf{1}_S) = \mathbf{0}_S$, where $\mathbf{1}_S$ is a matrix with each entry equals to 1 and $\mathbf{0}_S$ is defined similarly. Further notice that $(4\|\mathbf{\Psi}^*\|_2^2 + 1)\lambda u^{-1} \le 1/2$, we complete the proof. □

# B Improved Convergence Rate Using Sparsity Pattern

We develop an improved spectral norm convergence rate using sparsity pattern in this section. We collect the proof for Theorem 3.8 first and then give technical lemmas that are needed for the proof.

## B.1 Proof of Theorem 3.8

*Proof of Theorem 3.8.* Let us define $S^{(\ell)} = \big\{(i,j) : |\Psi^{(\ell)}_{ij} - \Psi^*_{ij}| \ge u\big\}$, where $u$ is introduced in (A.1). Let $S^{(0)} = \{(i,j) : |\Psi^*_{ij}| \ge u\} = S$. Then Lemma B.5 implies

$$\|\boldsymbol{\lambda}^{(\ell-1)}_{\mathcal{E}_\ell}\|_{\mathrm{F}} \le \lambda\|\mathrm{w}(|\mathbf{\Psi}^*_S| - u)\|_{\mathrm{F}} + \lambda\sqrt{|S^{(\ell-1)}\cap S|} + \lambda\sqrt{|\mathcal{E}_\ell/S|}$$

For any $(i,j) \in \mathcal{E}_\ell/S$, we must have $|\widehat{\Psi}_{ij}| = |\widehat{\Psi}_{ij} - \Psi^*_{ij}| > u$ and thus $(i,j) \in S^{(\ell-1)}/S$. Therefore, applying Lemma B.5 and using the fact that $\|\mathbf{\Psi}^*_S\|_{\max} \ge u + \alpha\lambda$, we obtain

$$\|\widehat{\mathbf{\Psi}}^{(\ell)} - \widehat{\mathbf{\Psi}}^\circ\|_{\mathrm{F}} \le 32\|\mathbf{\Psi}^*\|_2^2\lambda\Big\{\sqrt{|S^{(\ell-1)}\cap S|} + \sqrt{|S^{(\ell-1)}/S|}\Big\} \le 32\sqrt{2}\|\mathbf{\Psi}^*\|_2^2\lambda\sqrt{S^{(\ell-1)}}.$$

On the other side, $(i,j) \in S^{(\ell)}$ implies that

$$|\widehat{\Psi}^{(\ell)}_{ij} - \widehat{\Psi}^\circ_{ij}| \ge |\widehat{\Psi}^{(\ell)}_{ij} - \Psi^*_{ij}| - |\widehat{\Psi}^\circ_{ij} - \Psi^*_{ij}| \ge u - 2\kappa_2\lambda \ge 64\|\mathbf{\Psi}^*\|_2^2\lambda,$$

Exploiting the above fact, we can bound $\sqrt{|S^{(\ell)}|}$ in terms of $\|\widehat{\mathbf{\Psi}}^{(\ell)} - \widehat{\mathbf{\Psi}}^\circ\|_{\mathrm{F}}$:

$$\sqrt{|S^{(\ell)}|} \le \frac{\|\widehat{\mathbf{\Psi}}^{(\ell)} - \widehat{\mathbf{\Psi}}^\circ\|_{\mathrm{F}}}{64\|\mathbf{\Psi}^*\|_2^2\lambda} \le \sqrt{|S^{(\ell-1)}|/2}.$$

By induction on $\ell$, we obtain

$$\sqrt{|S^{(\ell)}|} \le \Big(\frac{1}{2}\Big)^{\ell/2}\sqrt{|S^{(0)}|} = \Big(\frac{1}{2}\Big)^{\ell/2}\sqrt{s}.$$

Since $\ell > \log s/\log 2$, we must have that the right hand side of the above inequality is smaller than 1, which implies that

$$S^{(\ell)} = \varnothing \text{ and } \widehat{\boldsymbol{\Psi}}^{(\ell)} = \widehat{\boldsymbol{\Psi}}^{\circ}.$$

Therefore, the estimator enjoys the strong oracle property. Using Lemma B.4 obtains us that

$$\big\|\widehat{\boldsymbol{\Psi}}^{(\ell)} - \boldsymbol{\Psi}^*\big\|_2 \le \big\|\widehat{\boldsymbol{\Psi}}^{\circ} - \boldsymbol{\Psi}^*\big\|_2 \lesssim \|\mathbf{M}^*\|_2 \big\|(\widehat{\mathbf{C}} - \mathbf{C}^*)_S\big\|_{\max}.$$

Applying Lemma D.6 finishes the proof of theorem. □

## B.2 Technical Lemmas

We start with the definitions of some constants. For notational simplicity, let $\kappa_1 = \|\boldsymbol{\Sigma}^*\|_\infty$ and $\mathcal{D} = \{(i,i) : 1 \le i \le d\}$. Define the oracle estimator as

$$\widehat{\boldsymbol{\Psi}}^{\circ} = \mathop{\mathrm{argmin}}_{\mathrm{supp}(\boldsymbol{\Psi}) = S, \boldsymbol{\Psi} \in \mathcal{S}_+^d} \Big\{ \big\langle \boldsymbol{\Psi}, \widehat{\mathbf{C}} \big\rangle - \log\det(\boldsymbol{\Psi}) \Big\}.$$

Recall that $s_{\max} = \max_j \sum_i 1(\Theta^*_{ij})$ is the maximum degree.

**Lemma B.1.** Suppose that the weight function satisfies that $\mathrm{w}(u) \ge 1/2$ for $u$ defined in (A.1). Assume that $2\lambda s_{\max} \le \kappa_1^{-2}\|\boldsymbol{\Psi}^*\|_2$, $8\|\boldsymbol{\Psi}^*\|_2^2\lambda\sqrt{s} < 1$. If $\lambda \ge 2\|\nabla\mathcal{L}(\widehat{\boldsymbol{\Psi}}^{\circ})\|_{\max}$, we must have

$$|\mathcal{E}_\ell| \le 2s \quad \text{and} \quad \big\|\widehat{\boldsymbol{\Psi}}^{(\ell)} - \widehat{\boldsymbol{\Psi}}^{\circ}\big\|_{\mathrm{F}} \le 32\big\|\boldsymbol{\Psi}^*\big\|_2^2 \big\|\boldsymbol{\lambda}^{(\ell-1)}_{\mathcal{E}_\ell}\big\|_{\mathrm{F}}.$$

*Proof of Lemma B.1.* If we assume that for all $\ell \ge 1$, we have the following

$$|\mathcal{E}_\ell| \le 2s, \text{ where } \mathcal{E}_\ell \text{ is defined in (A.1), and} \tag{B.1}$$

$$\|\boldsymbol{\lambda}^{(\ell-1)}_{\mathcal{E}_\ell^c}\|_{\min} \ge \|\nabla\mathcal{L}(\widehat{\boldsymbol{\Psi}}^{\circ})\|_{\max}. \tag{B.2}$$

Using lemma B.4, we obtain that $\|\widehat{\boldsymbol{\Psi}}^{\circ}\|_2 \le \|\boldsymbol{\Psi}^*\|_2 + \|\widehat{\boldsymbol{\Psi}}^{\circ} - \boldsymbol{\Psi}^*\|_\infty \le \|\boldsymbol{\Psi}^*\|_2 + 2\kappa_2\lambda s_{\max}$. Therefore, the assumption of the lemma implies $4\|\widehat{\boldsymbol{\Psi}}^{\circ}\|_2\lambda\sqrt{s} < 1$. Replacing $S$ by $\mathcal{E}_\ell$ in Lemma B.3 and using Hölder inequality, we have

$$\big\|\widehat{\boldsymbol{\Psi}}^{(\ell)} - \widehat{\boldsymbol{\Psi}}^{\circ}\big\|_{\mathrm{F}} \le 4\big\|\widehat{\boldsymbol{\Psi}}^{\circ}\big\|_2^2\big\|\boldsymbol{\lambda}^{(\ell-1)}_{\mathcal{E}_\ell}\big\|_{\mathrm{F}} \le 16\big\|\boldsymbol{\Psi}^*\big\|_2^2\big\|\boldsymbol{\lambda}^{(\ell-1)}_{\mathcal{E}_\ell}\big\|_{\mathrm{F}} \le 32\big\|\boldsymbol{\Psi}^*\big\|_2^2\lambda\sqrt{s}, \tag{B.3}$$

For $\ell = 1$, we have $\lambda \ge \lambda\mathrm{w}(u)$ and thus $\mathcal{E}_1 = S$, which implies that (B.1) and (B.2) hold for $\ell = 1$. Now assume that (B.1) and (B.2) hold at $\ell - 1$ for some $\ell \ge 2$. Since $j \in \mathcal{E}_\ell \setminus S$ implies that $j \notin S$ and $\lambda\mathrm{w}(\beta_j^{(\ell-1)}) = \lambda_j^{(\ell)} < \lambda\mathrm{w}(u)$ by assumption, and since $\mathrm{w}(x)$ is decreasing, we must have $|\beta_j^{(\ell-1)}| \ge u$. Therefore by induction hypothesis, we obtain that

$$\sqrt{|\mathcal{E}_\ell \setminus S|} \le \frac{\big\|\widehat{\boldsymbol{\Psi}}^{(\ell-1)}_{\mathcal{E}_\ell\setminus S}\big\|_{\mathrm{F}}}{u} \le \frac{\big\|\widehat{\boldsymbol{\Psi}}^{(\ell-1)} - \widehat{\boldsymbol{\Psi}}^{\circ}\big\|_{\mathrm{F}}}{u} \le \frac{32\big\|\boldsymbol{\Psi}^*\big\|_2^2\lambda}{u}\sqrt{s} \le \sqrt{s},$$

where the last inequality follows from the definition of $u$ hold at $\ell - 1$. This inequality implies that $|\mathcal{E}_\ell| \le 2|S| = 2s$. Now for such $\mathcal{E}_\ell^c$, we have

$$\|\boldsymbol{\lambda}_{\mathcal{E}_\ell^c}\|_{\min} \ge \lambda\mathrm{w}(u) \ge \lambda/2 \ge \|\nabla\mathcal{L}(\widehat{\boldsymbol{\Psi}}^{\circ})\|_{\max},$$

which completes the induction step. This completes the proof. □

With some abuse of notation, we let $|\mathbf{\Psi}_S^*| = (|\Psi_{ij}^*|)_{(i,j)\in S}$ and $|\mathbf{\Psi}_S^*| - u = (\Psi_{ij}^* - u)_{(i,j)\in S}$. The following inequality bounds the regularization parameter $\boldsymbol{\lambda}_{\mathcal{E}} = \lambda \mathrm{w}(|\mathbf{\Psi}_{\mathcal{E}}^*|) = \big(\lambda \mathrm{w}(\Psi_{ij}^*)\big)_{(i,j)\in\mathcal{E}}$ in terms of functionals of $\mathbf{\Psi}^*$ and $\mathbf{\Psi}$.

**Lemma B.2.** Let $\boldsymbol{\lambda} = \lambda \mathrm{w}\big(|\mathbf{\Psi}|\big)$. For any set $\mathcal{E} \supseteq S$, $\boldsymbol{\lambda}_{\mathcal{E}}$ must satisfy

$$\big\|\boldsymbol{\lambda}_{\mathcal{E}}\big\|_{\mathrm{F}} \le \lambda \big\|\mathrm{w}(|\mathbf{\Psi}_S^*| - u)\big\|_{\mathrm{F}} + \lambda\sqrt{|\mathcal{E}/S|} + \lambda\big|\{j \in S : |\Psi_{ij} - \Psi_{ij}^*| \ge u\}\big|^{1/2}$$

*Proof.* By triangle inequality, we have $\|\boldsymbol{\lambda}_{\mathcal{E}}\|_{\mathrm{F}} \le \|\boldsymbol{\lambda}_S\|_{\mathrm{F}} + \lambda\sqrt{|\mathcal{E}/S|}$. We further bound $\|\boldsymbol{\lambda}_S\|_{\mathrm{F}}$. If $|\Psi_{ij} - \Psi_{ij}^*| \ge u$, then we have $\mathrm{w}\big(|\Psi_{ij}|\big) \le 1 \le I\big(|\Psi_{ij} - \Psi_{ij}^*| \ge u\big)$, otherwise, since because $\mathrm{w}(\cdot)$ is non-increasing and thus $|\Psi_{ij} - \Psi_{ij}^*| < u$ implies $\mathrm{w}\big(|\Psi_{ij}|\big) \le \mathrm{w}\big(|\Psi_{ij}^*| - u\big)$. Therefore, using the Cauchy Schwartz inequality completes our proof. ◻

Define the following optimization problem

$$\widehat{\mathbf{\Psi}} = \operatorname*{argmin}_{\mathbf{\Psi}\in\mathcal{S}_+^d} \Big\{\big\langle \mathbf{\Psi}, \widehat{\mathbf{C}}\big\rangle - \log\det(\mathbf{\Psi}) + \big\|\boldsymbol{\lambda}\odot\mathbf{\Psi}\big\|_{1,\mathrm{off}}\Big\}. \tag{B.4}$$

**Lemma B.3.** Let $\|\boldsymbol{\lambda}_{S^c/\mathcal{D}}\|_{\min} \ge \|\nabla\mathcal{L}(\widehat{\mathbf{\Psi}}^\circ)\|_{\max}$ and $4\|\widehat{\mathbf{\Psi}}^\circ\|_2\lambda\sqrt{s} < 1$. Then $\widehat{\mathbf{\Psi}}$ must satisfy

$$\big\|\widehat{\mathbf{\Psi}} - \widehat{\mathbf{\Psi}}^\circ\big\|_{\mathrm{F}} \le 4\big\|\widehat{\mathbf{\Psi}}^\circ\big\|_2^2\big\|\boldsymbol{\lambda}_S\big\|_{\mathrm{F}}.$$

*Proof.* We construct an intermediate solution $\widetilde{\mathbf{\Theta}} = \mathbf{\Theta}^* + t(\widehat{\mathbf{\Theta}} - \mathbf{\Theta}^*)$, where $t$ is chosen such that $\|(\widetilde{\mathbf{\Theta}} - \mathbf{\Theta}^*\|_{\mathrm{F}} = r$, if $\|(\widehat{\mathbf{\Theta}} - \mathbf{\Theta}^*\|_{\mathrm{F}} > r$; $t = 1$ otherwise. Here $r$ satisfies $4\|\widehat{\mathbf{\Psi}}^\circ\|_2^2\lambda\sqrt{s} < r \le \|\widehat{\mathbf{\Psi}}^\circ\|_2$. Lemma A.3 implies that

$$\big(\|\widehat{\mathbf{\Psi}}^\circ\|_2 + r\big)^{-2}\big\|\widetilde{\mathbf{\Psi}} - \widehat{\mathbf{\Psi}}^\circ\big\|_{\mathrm{F}} \le \big\langle\nabla\mathcal{L}(\widetilde{\mathbf{\Psi}}) - \nabla\mathcal{L}(\widehat{\mathbf{\Psi}}^\circ), \widetilde{\mathbf{\Psi}} - \widehat{\mathbf{\Psi}}^\circ\big\rangle \equiv D_{\mathcal{L}}^s\big(\widetilde{\mathbf{\Psi}}, \widehat{\mathbf{\Psi}}^\circ\big). \tag{B.5}$$

Then, we use Lemma E.2 to upper bound the right hand side of the above inequality

$$D_{\mathcal{L}}^s(\widetilde{\mathbf{\Psi}}, \widehat{\mathbf{\Psi}}^\circ) \le tD_{\mathcal{L}}^s(\widehat{\mathbf{\Psi}}, \widehat{\mathbf{\Psi}}^\circ) = t\big\langle\nabla\mathcal{L}(\widehat{\mathbf{\Psi}}) - \nabla\mathcal{L}(\widehat{\mathbf{\Psi}}^\circ), \widetilde{\mathbf{\Psi}} - \widehat{\mathbf{\Psi}}^\circ\big\rangle.$$

Plugging the above inequality into (B.5), we obtain

$$\big(\|\widehat{\mathbf{\Psi}}^\circ\|_2 + r\big)^{-2}\big\|\widetilde{\mathbf{\Psi}} - \widehat{\mathbf{\Psi}}\big\|_{\mathrm{F}}^2 \le \big\langle\nabla\mathcal{L}(\widehat{\mathbf{\Psi}}) - \nabla\mathcal{L}(\widehat{\mathbf{\Psi}}^\circ), \widetilde{\mathbf{\Psi}} - \widehat{\mathbf{\Psi}}^\circ\big\rangle. \tag{B.6}$$

We further control the right hand side of the above inequality by exploiting the first order optimality condition, which is $\nabla\mathcal{L}(\widehat{\mathbf{\Psi}}) + \boldsymbol{\lambda}\odot\widehat{\mathbf{\Gamma}} = \mathbf{0}$ and $\nabla\mathcal{L}(\widehat{\mathbf{\Psi}}^\circ)_{S\cup\mathcal{D}} = \mathbf{0}$. Therefore, adding and subtracting term $\boldsymbol{\lambda}\odot\widehat{\mathbf{\Gamma}}$ to the right hand side of (B.6) and using the optimality condition obtains us that

$$\big(\|\widehat{\mathbf{\Psi}}^\circ\|_2 + r\big)^{-2}\big\|\widetilde{\mathbf{\Psi}} - \widehat{\mathbf{\Psi}}^\circ\big\|_{\mathrm{F}}^2 + \underbrace{\big\langle\boldsymbol{\lambda}\odot\widehat{\mathbf{\Gamma}}, \widetilde{\mathbf{\Psi}} - \widehat{\mathbf{\Psi}}^\circ\big\rangle}_{\mathrm{I}} + \underbrace{\big\langle\nabla\mathcal{L}(\widehat{\mathbf{\Psi}}^\circ), \widetilde{\mathbf{\Psi}} - \widehat{\mathbf{\Psi}}^\circ\big\rangle}_{\mathrm{II}} \le 0. \tag{B.7}$$

Therefore, to bound $\|\widetilde{\mathbf{\Psi}} - \widehat{\mathbf{\Psi}}^\circ\|_{\mathrm{F}}^2$, it suffices to bound I and II separately. For term I, by decomposing the support to $S$ and $S^c/\mathcal{D}$, then using matrix Hölder inequality, we have

$$\mathrm{I} \ge -\big\|\boldsymbol{\lambda}_S\big\|_{\mathrm{F}}\big\|\big(\widetilde{\mathbf{\Psi}} - \widehat{\mathbf{\Psi}}^\circ\big)_S\big\|_{\mathrm{F}} + \big\|\boldsymbol{\lambda}_{S^c/\mathcal{D}}\big\|_{\min}\big\|\mathrm{vec}\big(\widetilde{\mathbf{\Psi}} - \widehat{\mathbf{\Psi}}\big)_{S^c/\mathcal{D}}\big\|_1.$$

Again, by using the optimality condition, we has

$$\text{II} = \big\langle \nabla\mathcal{L}(\widehat{\boldsymbol{\Psi}}^\circ)_{S^c/\mathcal{D}}, (\widetilde{\boldsymbol{\Psi}} - \widehat{\boldsymbol{\Psi}}^\circ)_{S^c/\mathcal{D}}\big\rangle \geq -\big\|\nabla\mathcal{L}(\widehat{\boldsymbol{\Psi}}^\circ)_{S^c/\mathcal{D}}\big\|_{\max}\big\|\text{vec}(\widetilde{\boldsymbol{\Psi}} - \widehat{\boldsymbol{\Psi}}^\circ)_{S^c/\mathcal{D}}\big\|_1.$$

By plugging the upper bound for I and II back into (B.7), we have

$$\begin{aligned}&\big(\|\widehat{\boldsymbol{\Psi}}^\circ\|_2 + r\big)^{-2}\big\|\widetilde{\boldsymbol{\Psi}} - \widehat{\boldsymbol{\Psi}}^\circ\big\|_{\text{F}}^2 + \big(\|\boldsymbol{\lambda}_{S^c/\mathcal{D}}\|_{\min} - \|\nabla\mathcal{L}(\widehat{\boldsymbol{\Psi}}^\circ)_{S^c/\mathcal{D}}\|_{\max}\big)\big\|\text{vec}(\widetilde{\boldsymbol{\Psi}} - \widehat{\boldsymbol{\Psi}})\big\|_1 \\ &\quad \leq \big\|\boldsymbol{\lambda}_S\big\|_{\text{F}}\big\|(\widetilde{\boldsymbol{\Psi}} - \widehat{\boldsymbol{\Psi}}^\circ)_S\big\|_{\text{F}}.\end{aligned}$$

By assumption, we know that $\|\boldsymbol{\lambda}\|_{\min} \geq \|\nabla\mathcal{L}(\widehat{\boldsymbol{\Psi}}^\circ)\|_{\max}$, which implies that the second term in the right hand side of the above inequality is positive. Thus, we have $\big(\|\widehat{\boldsymbol{\Psi}}^\circ\|_2 + r\big)^{-2}\big\|\widetilde{\boldsymbol{\Psi}} - \widehat{\boldsymbol{\Psi}}^\circ\big\|_{\text{F}} \leq \big\|\boldsymbol{\lambda}_S\big\|_{\text{F}}$. Now since $4\|\widehat{\boldsymbol{\Psi}}\|_2^2\lambda\sqrt{s} < r \leq \|\widehat{\boldsymbol{\Psi}}^\circ\|_2$, we obtain that $\big\|\widetilde{\boldsymbol{\Psi}} - \widehat{\boldsymbol{\Psi}}^\circ\big\|_{\text{F}} \leq 4\big\|\widehat{\boldsymbol{\Psi}}\big\|_2^2\big\|\boldsymbol{\lambda}_S\big\|_{\text{F}} \leq 4\|\widehat{\boldsymbol{\Psi}}^\circ\|_2^2\lambda\sqrt{s} < r$. By the construction of $\widetilde{\boldsymbol{\Psi}}$, we must have $t = 1$, and thus $\widetilde{\boldsymbol{\Psi}} = \widehat{\boldsymbol{\Psi}}$. □

Recall that $\mathbf{M}^*$ is the sparsity pattern matrix corresponding to $\boldsymbol{\Psi}^*$.

**Lemma B.4.** If $4\kappa_1^4 c_n + 1 < \sqrt{1+4\kappa_1/s_{\max}}$ and $\|(\widehat{\mathbf{C}} - \mathbf{C}^*)_S\|_{\max} \leq c_n/2$ for a sequence $c_n$, then we have

$$\big\|\widehat{\boldsymbol{\Psi}}^\circ - \boldsymbol{\Psi}^*\big\|_{\max} \leq \kappa_1^2 c_n \text{ and } \big\|\widehat{\boldsymbol{\Psi}}^\circ - \boldsymbol{\Psi}^*\big\|_2 \leq \kappa_1^2 c_n\|\mathbf{M}^*\|_2.$$

*Proof of Lemma B.4.* Let $\Delta = \widehat{\boldsymbol{\Psi}}^\circ - \boldsymbol{\Psi}^*$. It suffices to show that $\|\boldsymbol{\Delta}\|_{\max} \leq r$, where $r = \kappa_1^2 c_n$. To show this, we construct an intermediate estimator, $\widetilde{\boldsymbol{\Psi}} = \boldsymbol{\Psi}^* + t(\widehat{\boldsymbol{\Psi}}^\circ - \boldsymbol{\Psi}^*)$. We choose $t$ such that $\|\widetilde{\boldsymbol{\Psi}} - \boldsymbol{\Psi}^*\|_{\max} = r$, if $\|\boldsymbol{\Delta}\|_{\max} > r$, and $\widetilde{\boldsymbol{\Psi}} = \widehat{\boldsymbol{\Psi}}$, otherwise. For a matrix $\mathbf{A}$, let $\mathbf{A}_S$ be a matrix agreeing with $\mathbf{A}$ on $S$ and having 0 elsewhere. Using the two term Taylor expansion, we know that there exists a $\gamma \in [0, 1]$ such that $\widetilde{\boldsymbol{\Psi}}^* = \boldsymbol{\Psi}^* + \gamma(\widetilde{\boldsymbol{\Psi}} - \boldsymbol{\Psi}^*)$,

$$\text{vec}\big\{\nabla\mathcal{L}(\widetilde{\boldsymbol{\Psi}})\big\} = \text{vec}\big\{\nabla\mathcal{L}(\boldsymbol{\Psi}^*)\big\} + \nabla^2\mathcal{L}(\widetilde{\boldsymbol{\Psi}}^*)\text{vec}\big(\widetilde{\boldsymbol{\Psi}} - \boldsymbol{\Psi}^*\big),$$

which implies that

$$\text{vec}\Big\{\mathbf{C}_{\mathcal{E}}^* - (\widetilde{\boldsymbol{\Psi}})_{\mathcal{E}}^{-1}\Big\} - \Big(\widetilde{\boldsymbol{\Psi}}_{\mathcal{E}}^* \otimes \widetilde{\boldsymbol{\Psi}}_{\mathcal{E}}^*\Big)^{-1}\text{vec}\Big(\widetilde{\boldsymbol{\Psi}}_{\mathcal{E}} - \boldsymbol{\Psi}_{\mathcal{E}}^*\Big) = \mathbf{0}, \tag{B.8}$$

where $\mathcal{E} = S \cup \mathcal{D}$. Let $\widetilde{\Delta} = \widetilde{\boldsymbol{\Psi}}_{\mathcal{E}} - \boldsymbol{\Psi}_{\mathcal{E}}^* = t\boldsymbol{\Delta}$. Define $f\big(\text{vec}(\widetilde{\boldsymbol{\Delta}})\big)$ to be

$$\Big\|\text{vec}\Big\{\mathbf{C}_{\mathcal{E}}^* - (\boldsymbol{\Psi}^* + \widetilde{\boldsymbol{\Delta}})_{\mathcal{E}}^{-1}\Big\} - \boldsymbol{\Gamma}_{\mathcal{E}\mathcal{E}}^*\text{vec}(\widetilde{\boldsymbol{\Delta}}_{\mathcal{E}})\Big\|_\infty,$$

in which $\boldsymbol{\Gamma}_{\mathcal{E}\mathcal{E}}^* = (\boldsymbol{\Psi}_{\mathcal{E}}^* \otimes \boldsymbol{\Psi}_{\mathcal{E}}^*)^{-1}$. By the matrix expansion formula that $(\mathbf{A} + \boldsymbol{\Delta})^{-1} - \mathbf{A}^{-1} = \sum_{m=1}^\infty(-\mathbf{A}^{-1}\boldsymbol{\Delta})^m\mathbf{A}^{-1}$, $f\{\text{vec}(\widetilde{\boldsymbol{\Delta}})\}$ reduces to

$$\Bigg\|\text{vec}\Bigg[\Bigg\{\sum_{m=2}^\infty(-\boldsymbol{\Sigma}^*\widetilde{\boldsymbol{\Delta}})^m\boldsymbol{\Sigma}^*\Bigg\}_{\mathcal{E}}\Bigg]\Bigg\|_\infty.$$

Using triangle inequality, we then obtain that

$$f\big\{\text{vec}(\widetilde{\boldsymbol{\Delta}})\big\} \leq \max_{(j,k)\in\mathcal{E}}\sum_{m=2}^\infty\Big|\mathbf{e}_j^T(\boldsymbol{\Sigma}^*\widetilde{\boldsymbol{\Delta}})^m\boldsymbol{\Sigma}^*\mathbf{e}_k\Big|.$$

Further applying Hölder inequality to each single term in the right hand side of the above displayed inequality, we have

$$\left|\mathbf{e}_j^T(\mathbf{\Sigma}^*\widetilde{\mathbf{\Delta}})^m\mathbf{\Sigma}^*\mathbf{e}_k\right| \le \left\|\mathbf{\Sigma}^*\right\|_\infty^{m+1}\|\widetilde{\mathbf{\Delta}}\|_\infty^{m-1}\|\widetilde{\mathbf{\Delta}}\|_{\max} \le s_{\max}^{m-1}\left\|\mathbf{\Sigma}^*\right\|_\infty^{m+1}\|\widetilde{\mathbf{\Delta}}\|_{\max}^m,$$

where we use the fact $\|\mathbf{\Delta}\|_\infty \le s_{\max}\|\mathbf{\Delta}\|_{\max}$. Therefore, we obtain

$$f\left\{\mathrm{vec}(\widetilde{\mathbf{\Delta}})\right\} \le \sum_{m=2}^{\infty} s_{\max}^{m-1}\|\mathbf{\Sigma}^*\|_\infty^{m+1}\|\widetilde{\mathbf{\Delta}}\|_{\max}^m = \frac{\kappa_1^3 s_{\max}\|\widetilde{\mathbf{\Delta}}\|_{\max}^2}{1-\kappa_1 s_{\max}\|\widetilde{\mathbf{\Delta}}\|_{\max}},$$

which, by triangle inequality, implies that

$$\|\widetilde{\mathbf{\Delta}}\|_{\max} \le \|\mathbf{\Gamma}_{\mathcal{E}\mathcal{E}}^*\|_\infty\left(\left\|\mathrm{vec}\left\{\mathbf{C}_{\mathcal{E}}^* - (\mathbf{\Psi}^*+\widetilde{\mathbf{\Delta}})_{\mathcal{E}}^{-1}\right\}\right\|_\infty + \frac{\kappa_1^3 s_{\max}\|\widetilde{\mathbf{\Delta}}\|_{\max}^2}{1-\kappa_1 s_{\max}\|\widetilde{\mathbf{\Delta}}\|_{\max}}\right).$$

Utilizing the KKT condition $\widehat{\mathbf{C}}_{\mathcal{E}} = \widehat{\mathbf{\Psi}}_{\mathcal{E}}^\circ$, the fact $\|\widehat{\mathbf{C}} - \mathbf{C}^*\|_{\max} \le c_n/2$ and $4\kappa_1^4 c_n < -1+\sqrt{1+\kappa_1/s_{\max}}$, we obtain

$$\|\widetilde{\mathbf{\Delta}}\|_{\max} \le \kappa_1^2 c_n\left(\frac{1}{2} + \frac{\kappa_1^3 s_{\max} r^2}{1-\kappa_1 s_{\max} r}\right) < \kappa_1^2 c_n \equiv r,$$

which is a contradiction. Thus, $\widetilde{\mathbf{\Delta}} = \mathbf{\Delta}$ and $\widehat{\mathbf{\Psi}}^\circ$ satisfies the desired maximum norm bound. For the spectral norm bound, we utilize Lemma E.6 and obtain that

$$\left\|\widehat{\mathbf{\Psi}}^\circ - \mathbf{\Psi}^*\right\|_2 \le \left\|\mathbf{M}^*\right\|_2\left\|\widehat{\mathbf{\Psi}}^\circ - \mathbf{\Psi}^*\right\|_{\max} \le \kappa_1^2 c_n\|\mathbf{M}^*\|_2.$$

The proof is finished. □

# C Semiparametric Graphical Model

*Proof of Theorem 4.3.* We need the follows lemma, which are taken from Liu et al. (2012). It provides a nonasymptotic probability bound for estimating $\mathbf{\Sigma}^{\mathrm{npn}}$ using $\widehat{\boldsymbol{S}}^\tau$.

**Lemma C.1.** Let $C$ be a constant. For any $n \gtrsim \log d$, with probability at least $1-8/d$, we have

$$\sup_{jk}|\widehat{S}_{jk}^\tau - \Sigma_{jk}^{\mathrm{npn}}| \le C\sqrt{\frac{\log d}{n}}.$$

The rest of the proof is adapted from that of Theorem 4.3 and thus is omitted.

□

# D Concentration Inequality

In this section, we establish the concentration inequalities which are the key technical tools to the large probability bounds in Section 3.

**Lemma D.1** (Sub-Gaussian Tail Bound). Let $\boldsymbol{X} = (X_1, X_2, \ldots, X_d)^{\mathrm{T}}$ be a zero-mean random vector with covariance $\boldsymbol{\Sigma}^*$ such that each $X_i/\sigma^*_{ii}$ is sub-Gaussian with variance proxy 1. Then there exists constants $c_1$ and $t_0$ such that for all $t$ with $0 \le t \le t_0$ the associated sample covariance $\widehat{\boldsymbol{\Sigma}}$ satisfies the following tail probability bound

$$\mathbb{P}\big(|\widehat{\sigma}_{ij} - \sigma^*_{ij}| \ge t\big) \le 8\exp\big\{-c_1 n t^2\big\}.$$

*Proof of Lemma D.1.* By the definition of the sample covariance matrix, we have $\widehat{\sigma}_{ij} = n^{-1}\sum_{k=1}^n (X_i^{(k)} - \bar{X}_i)(X_j^{(k)} - \bar{X}_j) = n^{-1}\sum_{k=1}^n X_i^{(k)}X_j^{(k)} - \bar{X}_i\bar{X}_j$. Therefore we can decompose $\widehat{\sigma}_{ij} - \sigma^*_{ij}$ as $n^{-1}\sum_{k=1}^n X_i^{(k)}X_j^{(k)} - \sigma^*_{ij} - \bar{X}_i\bar{X}_j$. By applying the union sum bound, we obtain that

$$\mathbb{P}\bigg(\Big|\widehat{\sigma}_{ij} - \sigma^*_{ij}\Big| \ge t\bigg) \le \underbrace{\mathbb{P}\bigg(\Big|\frac{1}{n}\sum_{k=1}^n X_i^{(k)}X_j^{(k)} - \sigma^*_{ij}\Big| \ge \frac{t}{2}\bigg)}_{\text{(R1)}} + \underbrace{\mathbb{P}\bigg(\Big|\bar{X}_i\bar{X}_j\Big| \ge \frac{t}{2}\bigg)}_{\text{(R2)}}$$

In the sequel, we bound (R1) and (R2) separately. For term (R1), following the argument of Lemma A.3 in Bickel and Levina (2008), there exists constant $c_1'$ and $t_0'$ not depending $n, d$ such that

$$\text{(R1)} = \mathbb{P}\bigg(\Big|\frac{1}{n}\sum_{k=1}^n X_i^{(k)}X_j^{(k)} - \sigma^*_{ij}\Big| \ge \frac{t}{2}\bigg) \le 4\exp\Big\{-c_1' n t^2\Big\}$$

for all $t$ satisfying $0 \le t \le t_0$. Next, we bound the term (R2). By the linear structure of sub-Gaussian random variables, we obtain that $\sqrt{n}\bar{X}_i \sim \text{sub-Gaussian}(0, \sigma^*_{ii})$ for all $1 \le i \le d$. Therefore, by applying Lemma E.1, we obtain that $|\sqrt{n}\bar{X}_i \cdot \sqrt{n}\bar{X}_j|$ is a sub-exponential random variable with $\psi_1$ norm bounded by $2\|\sqrt{n}\bar{X}_i\|_{\psi_2}\|\sqrt{n}\bar{X}_j\|_{\psi_2}$. We give explicit bounds for the $\psi_2$-norm of $\sqrt{n}\bar{X}_i$ and $\sqrt{n}\bar{X}_j$. By the Chernoff bound, the tail probability of $\sqrt{n}\bar{X}_i$ can be bounded in the following

$$\mathbb{P}\Big(|\sqrt{n}\bar{X}_i| \ge t\Big) \le 2\exp\Big\{-\frac{t^2}{2\sigma^*_{ii}}\Big\}.$$

For every non-negative random variable $Z$, integration by parts yields the identity $\mathbb{E}Z = \int_0^\infty \mathbb{P}(Z \ge u)du$. We apply this for $Z = |\sqrt{n}\bar{X}_i|^p$ and obtain after change of variables $u = t^p$ that

$$\begin{aligned}\mathbb{E}|\sqrt{n}\bar{X}_i|^p &= \int_0^\infty \mathbb{P}(|\sqrt{n}\bar{X}_i| \ge t)\cdot pt^{p-1}dt \le \int_0^\infty 2p\cdot\exp\Big\{-\frac{t^2}{2\sigma^*_{ii}}\Big\}t^{p-1}dt \\ &= p(2\sigma^*_{ii})^{p/2}\cdot\Gamma(\frac{p}{2}) \le p(2\sigma^*_{ii})^{p/2}\cdot\Big(\frac{p}{2}\Big)^{p/2},\end{aligned}$$

which indicates that $\|\sqrt{n}\bar{X}_i\|_{\psi_1} \le \sqrt{2\sigma^*_{ii}}$. The Gamma function is defined as $\Gamma(t) = \int_0^\infty e^{-t}x^{t-1}dx$. Similary, we can bound $\|\sqrt{n}\ \bar{X}_j\|_{\psi_2}$ by $\sqrt{2\sigma^*_{jj}}$. Therefore we obtain $\|\sqrt{n}\bar{X}_i\cdot\sqrt{n}\bar{X}_j\|_{\psi_1} \le 2\sqrt{\sigma^*_{ii}\sigma^*_{jj}} \le 2\sigma^2_{\max}$, where $\sigma^2_{\max} = \max\{\sigma^*_{11}, \ldots, \sigma^*_{dd}\}$. Define $Z_{ij} = |\sqrt{n}\bar{X}_i\cdot\sqrt{n}\bar{X}_j|$. Let $\delta = (e-1)(2\sigma^2_{\max}e^2)^{-1}$ and write the Taylor expansion series of the expoential function, we obtain

$$\mathbb{E}\exp\{\delta Z_{ij}\} = 1 + \sum_{k=1}^\infty \frac{\delta^k\mathbb{E}(Z_{ij}^k)}{k!} \le 1 + \sum_{k=1}^\infty \frac{\delta^k(2\sigma^2_{\max}k)^k}{k!} \le 1 + \sum_{k=1}^\infty (2\sigma^2_{\max}\delta\cdot e)^k \le e,$$

where we use $k! \geq (k/e)^k$ in the last second inequality. Exponenting and using the Markov inequalty yields that

$$\mathbb{P}\Big(Z_{ij} \geq t\Big) = \mathbb{P}\Big(\delta Z_{ij} \geq \delta t\Big) = \mathbb{P}\Big(e^{\delta Z_{ij}} \geq e^{\delta t}\Big) \leq \frac{\mathbb{E}e^{\delta Z_{ij}}}{e^{\delta t}} \leq \exp\{1 - \delta t\},$$

for all $t \geq 0$. Using the above result, we can boudn (R2) as

$$(\text{R2}) \leq \mathbb{P}\Big(Z_{ij} \geq \frac{nt}{2}\Big) \leq \exp\Big\{1 - \frac{\delta nt}{2}\Big\} \leq 4\exp\Big\{1 - \frac{\delta nt}{2}\Big\}.$$

Combing the bounds for (R1) and (R2), taking $c_1 = \min\{c_1', \delta\}$ and $t_0 = \min\{1, t_0'\}$ obtain us that

$$\mathbb{P}\big(|\widehat{\sigma}_{ij} - \sigma_{ij}^*| \geq t\big) \leq 8\exp\big\{ - c_1 nt^2\big\} \;\forall\, t \leq t_0,$$

which completes the proof. □

We then develop a large deviation bound for marginal variances.

**Lemma D.2** (Large Deviation Bound for Marginal Variance)**.** Let $\boldsymbol{X} = (X_1, X_2, \ldots, X_d)^{\mathrm{T}}$ be a zero-mean random vector with covariance $\boldsymbol{\Sigma}^*$ such that each $X_i/\sqrt{\Sigma_{ii}^*}$ is sub-Gaussian with variance proxy 1, and $\big\{\boldsymbol{X}^{(k)}\big\}_{k=1}^n$ be $n$ i.i.d. samples from $\boldsymbol{X}$. Let $C(\varepsilon) = 2^{-1}\big(\varepsilon - \log(1+\varepsilon)\big) > 0$. Then, for any $\varepsilon \geq 0$, we must have

$$\mathbb{P}\Big(\big|\widehat{\Sigma}_{ii} - \Sigma_{ii}^*\big| > \varepsilon \cdot \Sigma_{ii}^*\Big) \leq 2 \cdot \exp\Big\{-n \cdot C(\varepsilon)\Big\}.$$

*Proof.* We write $Z_i^{(k)} = \big(\Sigma_{ii}^*\big)^{-1/2} X_i^{(k)}$ and $\widetilde{\Sigma}_{ii} = n^{-1}\sum_{k=1}^n Z_i^{(k)} \cdot Z_i^{(k)}$, for $1 \leq i \leq d$. Let $\varsigma_i^{(k)} = Z_i^{(k)} \cdot Z_i^{(k)} \sim \chi_1^2$, for $1 \leq k \leq n$. Therefore, the moment-generating function of $\varsigma_i^{(k)}$ is $M_{\varsigma_i^{(k)}}(t) = (1-2t)^{-1/2}$, for $t \in (-\infty, 1/2)$. Next, we control the tail probability of $\widetilde{\Sigma}_{ii} > 1+\varepsilon$ and $\widetilde{\Sigma}_{ii} < 1-\varepsilon$, respectively. For the tail probability of $\widetilde{\Sigma}_{ii} > 1+\varepsilon$, by applying Lemma E.8, we obtain

$$\mathbb{P}\left(\frac{\varsigma_i^{(1)} + \ldots + \varsigma_i^{(n)}}{n} > 1+\varepsilon\right) \leq \exp\Big\{-n \cdot A(\varepsilon)\Big\},$$

where $A(\varepsilon) = \sup_t \big\{(1+\varepsilon)t + 2^{-1}\log(1-2t)\big\} = 2^{-1}\big(\varepsilon - \log(1+\varepsilon)\big)$. Similarly, for any $\varepsilon > 0$, we obtain the tail probability of $\widetilde{\Sigma}_{ii} < 1-\varepsilon$ as

$$\mathbb{P}\left(\frac{\varsigma_i^{(1)} + \ldots + \varsigma_i^{(n)}}{n} < 1-\varepsilon\right) \leq \exp\Big\{-n \cdot B(\varepsilon)\Big\},$$

where $B(\varepsilon) = \sup_t \big\{(1-\varepsilon)t + 2^{-1}\log(1-2t)\big\}$. After some algebra, we obtain $B(\varepsilon) = -2^{-1}\big(\varepsilon + \log(1-\varepsilon)\big)$, if $\varepsilon < 1$; $B(\varepsilon) = +\infty$, otherwise. Let $C(\varepsilon) = \min\big\{A(\varepsilon), B(\varepsilon)\big\} = 2^{-1}\big(\varepsilon - \log(1+\varepsilon)\big)$. Therefore, combing the above two inequalities by union bound, we obtain $\mathbb{P}\Big(\big|n^{-1}\big(\varsigma_i^{(1)} + \ldots + \varsigma_i^{(n)}\big) - 1\big| > \varepsilon\Big) \leq 2 \cdot \exp\Big\{-n \cdot C(\varepsilon)\Big\}$. Note that we have $\widehat{\Sigma}_{ii} = (\Sigma_{ii}^*)^{-1} \cdot \widetilde{\Sigma}_{ii} = n^{-1}\big(\varsigma_{ii}^{(1)} + \ldots + \varsigma_{ii}^{(n)}\big)$. Thus, we obtain

$$\mathbb{P}\Big(\big|\widehat{\Sigma}_{ii} - \Sigma_{ii}^*\big| > \varepsilon \cdot \Sigma_{ii}^*\Big) \leq 2 \cdot \exp\Big\{-n \cdot C(\varepsilon)\Big\}.$$

□

Our next results characterizes a large deviation bound for sample correlation matrix.

**Lemma D.3** (Large Deviation Bound for Sample Correlation). Let $\boldsymbol{X} = (X_1, X_2, \ldots, X_d)^{\mathrm{T}}$ be a zero-mean random vector with covariance matrix $\boldsymbol{\Sigma}^*$ such that each $X_i/\sqrt{\Sigma^*_{ii}}$ is sub-Gaussian with variance proxy 1 and $\{\boldsymbol{X}^{(k)}\}_{k=1}^n$ be $n$ independent and identically distributed copies of $\boldsymbol{X}$. Let $\widehat{\boldsymbol{\Sigma}} = 1/n\sum_{k=1}^n \boldsymbol{X}^{(k)}{\boldsymbol{X}^{(k)}}^{\mathrm{T}}$ denote the sample covariance and $\widehat{\mathbf{C}} = \widehat{\mathbf{W}}^{-1}\widehat{\boldsymbol{\Sigma}}\widehat{\mathbf{W}}^{-1}$ denote the sample correlation matrix, where $\widehat{\mathbf{W}}^2$ is the diagonal matrix with diagonal elements of $\widehat{\boldsymbol{\Sigma}}$. Further let $\widehat{\rho}_{ij}$ and $\rho_{ij}$ be the $(i,j)$th element of $\widehat{\mathbf{C}}$ and $\mathbf{C}^*$ respectively. Define $c_2 = \min\{4^{-1}c_1 \min(\Sigma^*_{ii})^2, 1/6\}$. Then, for $0 \le \varepsilon \le \min\{1/2, t_0 \max_i \Sigma^*_{ii}\}$, we have

$$\mathbb{P}\Big(|\widehat{\rho}_{ij} - \rho_{ij}| > \varepsilon\Big) \le 6\exp\Big\{-c_2 n\cdot\varepsilon^2\Big\}, \quad \text{where } \ 1 \le i \ne j \le d.$$

*Proof of Lemma D.3.* We denote the sample correlation as $\widehat{\rho}_{ij} = (\widehat{\Sigma}_{ii}\cdot\widehat{\Sigma}_{jj})^{-1/2}\widehat{\Sigma}_{ij}$. To prove the tail probability bound. It suffices to prove the tail probability bound for $\widehat{\rho}_{ij} - \rho_{ij} > \varepsilon$ and $\widehat{\rho}_{ij} - \rho_{ij} < -\varepsilon$, respectively. We start with the tail probability bound for $\widehat{\rho}_{ij} - \rho_{ij} > \varepsilon$. Let us assume that $\rho_{ij} \ge 0$. Using the basic probability argument, we have $\mathbb{P}(A) = \mathbb{P}(A\cap B) + \mathbb{P}(A\cap B^c) \le \mathbb{P}(A) + \mathbb{P}(B^c)$. Thus, for any $0 \le t \le 1$ we obtain

$$\begin{aligned}
\mathbb{P}\Big(\widehat{\rho}_{ij} - \rho_{ij} > \varepsilon\Big) &= \mathbb{P}\Big(\widehat{\Sigma}_{ij} - (\widehat{\Sigma}_{ii}\widehat{\Sigma}_{jj})^{-1/2}\cdot\rho_{ij} > (\widehat{\Sigma}_{ii}\widehat{\Sigma}_{jj})^{-1/2}\cdot\varepsilon\Big) \\
&\le \underbrace{\mathbb{P}\Big(\widehat{\Sigma}_{ij} - (\Sigma^*_{ii}\Sigma^*_{jj})^{-1/2}(1-t)^{-1}\cdot\rho_{ij} > (\Sigma^*_{ii}\Sigma^*_{jj})^{-1/2}(1-t)^{-1}\cdot\varepsilon\Big)}_{\text{(R1.1)}} \\
&\quad + \mathbb{P}\Big(\widehat{\Sigma}_{ii} - \Sigma^*_{ii} > \Sigma^*_{ii}\cdot t\Big) + \mathbb{P}\Big(\widehat{\Sigma}_{jj} - \Sigma^*_{jj} > \Sigma^*_{jj}\cdot t\Big).
\end{aligned} \tag{D.1}$$

Next, we bound the term (R1.1). After some simple algebra, (R1.1) can be bounded by

$$\begin{aligned}
&\mathbb{P}\Big(\widehat{\Sigma}_{ij} - \Sigma^*_{ij} > (\varepsilon + \rho_{ij})\cdot(\Sigma^*_{ii}\Sigma^*_{jj})^{-1/2}(1-t)^{-1} - \Sigma^*_{ij}\Big) \\
&\le \mathbb{P}\Big(\widehat{\Sigma}_{ij} - \Sigma^*_{ij} > \varepsilon\big(\Sigma^*_{ii}\Sigma^*_{jj}\big)^{-1/2}(1+t) + t\cdot\Sigma^*_{ij}\Big)
\end{aligned}$$

Let $c_2' = c_1 \min_i(\Sigma^*_{ii})^2$, where $c_1$ is defined in Lemma D.1. If we apply Lemma D.1 with a better constant and Lemma D.2, then for any $0 \le \varepsilon \le t_0\sqrt{\Sigma^*_{ii}\Sigma^*_{jj}}$, in which $t_0$ is defined in Lemma D.1, we must have

$$\begin{aligned}
\mathbb{P}\Big(\widehat{\rho}_{ij} - \rho_{ij} > \varepsilon\Big) &\le \mathbb{P}\Big(\widehat{\Sigma}_{ij} - \Sigma^*_{ij} > \varepsilon\big(\Sigma^*_{ii}\Sigma^*_{jj}\big)^{-1/2}\Big) + \mathbb{P}\Big(\widehat{\Sigma}_{ii} - \Sigma^*_{ii} > t\cdot\Sigma^*_{ii}\Big) \\
&\quad + \mathbb{P}\Big(\widehat{\Sigma}_{jj} - \Sigma^*_{jj} > t\cdot\Sigma^*_{jj}\Big) \\
&\le 4\exp\Big\{-c_2' n\cdot\varepsilon^2\Big\} + 2\exp\Big\{-n\cdot\frac{1}{2}\big(t - \log(1+t)\big)\Big\}.
\end{aligned}$$

Let $c_2'' = \min\big\{c_2', 1/6\big\}$. Further, for any $0 \le \varepsilon \le \min\{1/2, t_0\max_i \Sigma^*_{ii}\}$, by taking $t = \varepsilon$ and using the inequality $t - \log(1+t) \ge 1/3\cdot t^2$ for all $t$ such that $0 \le t \le 1/2$, we obtain

$$\mathbb{P}\Big(\widehat{\rho}_{ij} - \rho_{ij} > \varepsilon\Big) \le 4\exp\Big\{-c_2'\varepsilon^2\cdot n\Big\} + 2\exp\Big\{-\frac{1}{6}\varepsilon^2\cdot n\Big\} \le 6\exp\Big\{-c_2'' n\cdot\varepsilon^2\Big\}.$$

If $\rho_{ij} < 0$, in the a similar fashion as before, we can obtain the the following tail probability bound

$$\mathbb{P}\Big(\widehat{\rho}_{ij} - \rho_{ij} > \varepsilon\Big) \leq \underbrace{\mathbb{P}\Big(\widehat{\Sigma}_{ij} - \Sigma^*_{ij} > \varepsilon\big(\Sigma^*_{ii}\Sigma^*_{jj}\big)^{-1/2} + \Sigma^*_{ij}\cdot(t^2 - t) - \varepsilon\sqrt{\Sigma^*_{ii}\Sigma^*_{jj}}\cdot t\Big)}_{\text{(R1.2)}}$$
$$+ \mathbb{P}\Big(\widehat{\Sigma}_{ii} - \Sigma^*_{ii} > t\cdot\Sigma^*_{ii}\Big) + \mathbb{P}\Big(\widehat{\Sigma}_{jj} - \Sigma^*_{jj} > t\cdot\Sigma^*_{jj}\Big).$$

To continue, we bound the term (R1.2) in the next. If take $t = \varepsilon \leq \min\big\{1/2, t_0 \max_i \Sigma^*_{ii}\big\} \leq 1/2 + 1/2|\rho_{ij}|$, we obtain that $\Sigma^*_{ij}\cdot(t^2 - t) - \varepsilon\sqrt{\Sigma^*_{ii}\Sigma^*_{jj}}\cdot t \geq -1/2\sqrt{\Sigma^*_{ii}\Sigma^*_{jj}}\cdot t$. Thus, we have

$$\begin{aligned}
\mathbb{P}\Big(\widehat{\rho}_{ij} - \rho_{ij} > \varepsilon\Big) &\leq \mathbb{P}\Big(\widehat{\Sigma}_{ij} - \Sigma^*_{ij} > \frac{1}{2}\varepsilon\big(\Sigma^*_{ii}\Sigma^*_{jj}\big)^{-1/2}\Big) + \mathbb{P}\Big(\widehat{\Sigma}_{ii} - \Sigma^*_{ii} > t\cdot\Sigma^*_{ii}\Big) \\
&\quad + \mathbb{P}\Big(\widehat{\Sigma}_{jj} - \Sigma^*_{jj} > t\cdot\Sigma^*_{jj}\Big) \\
&\leq 4\exp\Big\{-\frac{1}{4}c_2' n\cdot\varepsilon^2\Big\} + 2\exp\Big\{-\frac{1}{2}n\cdot\big(\varepsilon - \log(1+\varepsilon)\big)\Big\} \\
&\leq 6\exp\Big\{-c_2 n\cdot\varepsilon^2\Big\},
\end{aligned}$$

where $c_2 = \min\{4^{-1}c_2', 1/6\} = \min\{4^{-1}c_1\min(\Sigma^*_{ii})^2, 1/6\} \leq c_2''$. By combining above two cases, for $0 \leq \varepsilon \leq \min\{1/2, t_0\max_i \Sigma^*_{ii}\}$, we have $\mathbb{P}(\widehat{\rho}_{ij} - \rho_{ij} > \varepsilon) \leq 6\exp\{-c_2 n\cdot\varepsilon^2\}$. In a similar fashion, we obtain the same tail probability bound for $\widehat{\rho}_{ij} - \rho_{ij} < \varepsilon$, for $0 \leq \varepsilon \leq \min\{1/2, t_0\max_i \Sigma^*_{ii}\}$. Thus the proof is completed. □

**Lemma D.4.** Under the same conditions in Lemma (D.3). We have the following result hold

$$\lim_{M\to\infty}\limsup_n \mathbb{P}\bigg(\big\|\nabla\mathcal{L}(\boldsymbol{\Psi}^*)_S\big\|_{\max} > M\sqrt{\frac{1}{n}}\bigg) = 0, \quad \text{and} \quad \|\nabla\mathcal{L}(\boldsymbol{\Psi}^*)_S\|_{\mathrm{F}} = \mathcal{O}_{\mathbb{P}}\bigg(\sqrt{\frac{s}{n}}\bigg).$$

*Proof of Lemma D.4.* It is easy to check that $\big\|\nabla\mathcal{L}(\boldsymbol{\Psi}^*)_S\big\|_{\mathrm{F}} = \big\|\big(\widehat{\mathbf{C}} - \mathbf{C}^*\big)_S\big\|_{\mathrm{F}}$. By applying Lemma D.3 and the union sum bound, for any $M$ such that $0 \leq M \leq \min\big\{1/2, t_0\max_i \Sigma^*_{ii}\big\}\cdot\sqrt{n}$, in which $t_0$ is defined in Lemma D.3, we obtain

$$\mathbb{P}\bigg(\big\|\nabla\mathcal{L}(\boldsymbol{\Psi}^*)_S\big\|_{\max} > M\sqrt{\frac{1}{n}}\bigg) \leq s\cdot\exp\big\{-c_2M^2\big\} \leq \exp\big\{-c_2M^2 + \log s\big\}.$$

Taking $M$ such that $\sqrt{2c_2^{-1}\log s} \leq M \leq \min\big\{1/2, t_0\max_i \Sigma^*_{ii}\big\}\cdot\sqrt{n}$ and $M \to \infty$ in the above inequality obtains us that

$$\lim_{M\to\infty}\limsup_n \mathbb{P}\bigg(\big\|\nabla\mathcal{L}(\boldsymbol{\Psi}^*)_S\big\|_{\max} > M\sqrt{\frac{1}{n}}\bigg) = 0,$$

which implies that $\big\|\nabla\mathcal{L}(\boldsymbol{\Psi}^*)_S\big\|_{\mathrm{F}} = \mathcal{O}_{\mathbb{P}}(\sqrt{s/n})$. □

**Lemma D.5** (A Concentration Inequality for Sample Correlation Matrix)**.** Let $\widehat{\mathbf{C}}$, $\mathbf{C}^*$, $\widehat{\rho}_{ij}$ and $\rho^*_{ij}$ be defined in Lemma D.3. Suppose $n \geq 3\big(c_2t_1^2\big)^{-1}\cdot\log d$. Take $\lambda = \sqrt{3c_2^{-1}\cdot(\log d)/n} \asymp \sqrt{\log(d)/n}$, in which $c_2$ is defined as in Lemma D.3. Then $\widehat{\mathbf{C}}$ must satisfy

$$\mathbb{P}\Big(\|\widehat{\mathbf{C}} - \mathbf{C}^*\|_{\max} \leq \lambda\Big) \leq 1 - 8/d.$$

*Proof.* It is easy to check that $\nabla\mathcal{L}(\mathbf{C}^*) = \widehat{\mathbf{C}} - \mathbf{C}^*$. Therefore, applying Lemma D.3 and union sum bound, we obtain that, for any $\lambda \leq t_1 \equiv \min\{1/2, t_0 \max_i\{\Sigma^*_{ii}\}\}$ with $t_0$ defined in Lemma D.1,

$$\mathbb{P}\Big(\big\|\widehat{\mathbf{C}} - \mathbf{C}^*\big\|_{\max} > \lambda\Big) \leq 6d^2 \cdot \exp\{-c_2 n\lambda^2\}.$$

where $c_2 = \min\{4^{-1}c_1 \min(\Sigma^*_{ii})^2, 1/6\}$, in which $c_1$ is defined in Lemma D.1. , for $n$ sufficiently large such that $n \geq 3\big(c_2 t_1^2\big)^{-1} \cdot \log d$, by taking $\lambda = \sqrt{3c_2^{-1} \cdot (\log d)/n} \leq t_1$, we obtain $\mathbb{P}\big(\|\widehat{\mathbf{C}} - \mathbf{C}^*\|_{\max} \leq \lambda\big) = 1 - \mathbb{P}\big(\|\widehat{\mathbf{C}} - \mathbf{C}^*\|_{\max} > \lambda\big) \geq 1 - 6d^2 \cdot \exp\{-c_2 n\lambda^2\} \geq 1 - 8/d$. The proof is completed. ∎

**Lemma D.6.** Under the same conditions in Lemma D.5, we have

$$\lim_{M\to\infty} \limsup_{n} \mathbb{P}\bigg(\big\|\big(\widehat{\mathbf{C}} - \mathbf{C}^*\big)_S\big\|_{\max} > M\sqrt{\frac{1}{n}}\bigg) = 0, \quad \text{and} \quad \big\|\big(\widehat{\mathbf{C}} - \mathbf{C}^*\big)_S\big\|_{\max} = \mathcal{O}_{\mathbb{P}}\bigg(\sqrt{\frac{1}{n}}\bigg).$$

*Proof of Lemma D.6.* The proof is similar to that of Lemma D.5 and thus is omitted. ∎

# E Preliminary Lemmas

In this section we state and prove the technical lemmas used in previous sections. The following lemma establishes the tail bound type of the product of two sub-Gaussian random variables. Let $\|\cdot\|_{\psi_1}$ and $\|\cdot\|_{\psi_2}$ be the $\psi_1$- and $\psi_2$-norm defined in Vershynin (2010).

**Lemma E.1.** For $X$ and $Y$ being two sub-Gaussian random variables, then the absolute value of their product $|X \cdot Y|$ is a sub-exponential random variable with

$$\|X \cdot Y\|_{\psi_1} \leq 2 \cdot \|X\|_{\psi_2}\|Y\|_{\psi_2}.$$

*Proof of Lemma E.1.* To show $X \cdot Y$ is sub-exponential, it suffices to prove that the $\psi_1$-norm of $X \cdot Y$ is bounded. By the definition of the $\psi_1$-norm, we have

$$\|X \cdot Y\|_{\psi_1} = \sup_{p\geq 1} p^{-1}\big[\mathbb{E}|X \cdot Y|^p\big]^{1/p}. \tag{E.1}$$

We need to use the Hölder inequality as follows

$$\mathbb{E}\big[|\langle f, g\rangle|\big] \leq \big[\mathbb{E}|f|^r\big]^{1/r}\big[\mathbb{E}|g|^s\big]^{1/s}, \qquad \frac{1}{r} + \frac{1}{s} = 1,$$

where $f$ and $g$ are two random functions. If we choose $f = X^p$, $g = Y^p$ and $r = s = 2$ in the Hölder inequality, then the right hand side of (E.1) can be bounded by

$$\begin{aligned}
&\sup_{p\geq 1}\Big\{p^{-1}\big[\mathbb{E}|X|^{2p}\big]^{1/(2p)}\big[\mathbb{E}|Y|^{2p}\big]^{1/(2p)}\Big\}\\
\leq& 2\sup_{p\geq 1}\Big\{(2p)^{-1/2}\big[\mathbb{E}|X|^{2p}\big]^{1/(2p)}\Big\} \cdot \sup_{p\geq 1}\Big\{(2p)^{-1/2}\big[\mathbb{E}|Y|^{2p}\big]^{1/(2p)}\Big\}.
\end{aligned}$$

Therefore we obtain that $\|X \cdot Y\|_{\psi_1} \leq 2\|X\|_{\psi_2}\|Y\|_{\psi_2} < \infty$. The proof is completed. ∎

**Lemma E.2.** Let $D_{\mathcal{L}}(\boldsymbol{\Theta}_1, \boldsymbol{\Theta}_2) = \mathcal{L}(\boldsymbol{\Theta}_1) - \mathcal{L}(\boldsymbol{\Theta}_2) - \big\langle \mathcal{L}(\boldsymbol{\Theta}_2), \boldsymbol{\Theta}_1 - \boldsymbol{\Theta}_2 \big\rangle$ and $D^s_{\mathcal{L}}(\boldsymbol{\Theta}_1, \boldsymbol{\Theta}_2) = D_{\mathcal{L}}(\boldsymbol{\Theta}_1, \boldsymbol{\Theta}_2) + D_{\mathcal{L}}(\boldsymbol{\Theta}_2, \boldsymbol{\Theta}_1)$. For $\boldsymbol{\Theta}(t) = \boldsymbol{\Theta}^* + t(\boldsymbol{\Theta} - \boldsymbol{\Theta}^*)$ with $t \in (0, 1]$, we have that

$$D^s_{\mathcal{L}}(\boldsymbol{\Theta}(t), \boldsymbol{\Theta}^*) \leq t D^s_{\mathcal{L}}(\boldsymbol{\Theta}, \boldsymbol{\Theta}^*).$$

*Proof of Lemma E.2.* Let $Q(t) = D_{\mathcal{L}}(\boldsymbol{\Theta}(t), \boldsymbol{\Theta}^*) = \mathcal{L}(\boldsymbol{\Theta}(t)) - \mathcal{L}(\boldsymbol{\Theta}^*) - \big\langle \nabla\mathcal{L}(\boldsymbol{\Theta}^*), \boldsymbol{\Theta}(t) - \boldsymbol{\Theta}^* \big\rangle$. Since the derivative of $\mathcal{L}(\boldsymbol{\Theta}(t))$ with respect to $t$ is $\langle \nabla\mathcal{L}(\boldsymbol{\Theta}(t)), \boldsymbol{\Theta} - \boldsymbol{\Theta}^* \rangle$, then the derivative of $Q(t)$ is

$$Q'(t) = \big\langle \nabla\mathcal{L}(\boldsymbol{\Theta}(t)) - \nabla\mathcal{L}(\boldsymbol{\Theta}^*), \boldsymbol{\Theta} - \boldsymbol{\Theta}^* \big\rangle.$$

Therefore the Bregman divergence $D^s_{\mathcal{L}}(\boldsymbol{\Theta}(t) - \boldsymbol{\Theta}^*)$ can written as

$$D^s_{\mathcal{L}}(\widetilde{\boldsymbol{\Theta}}(t) - \boldsymbol{\Theta}^*) = \big\langle \nabla\mathcal{L}(\widetilde{\boldsymbol{\Theta}}(t)) - \nabla\mathcal{L}(\boldsymbol{\Theta}^*), t(\boldsymbol{\Theta} - \boldsymbol{\Theta}^*) \big\rangle = tQ'(t) \quad \text{for} \quad 0 < t \leq 1.$$

By plugging $t = 1$ in the above function equation, we have $Q'(1) = D^s_{\mathcal{L}}(\boldsymbol{\Theta}, \boldsymbol{\Theta}^*)$ as a special case. If we assume that $Q(t)$ is convex, then $Q'(t)$ is non-decreasing and thus

$$D^s_{\mathcal{L}}(\boldsymbol{\Theta}(t), \boldsymbol{\Theta}^*) = tQ'(t) \leq tQ'(1) = tD^s_{\mathcal{L}}(\boldsymbol{\Theta}, \boldsymbol{\Theta}^*).$$

Therefore the proof is completed. It remains to prove that $Q(t)$ is a convex function, i.e.

$$Q(\alpha_1 t_1 + \alpha_2 t_2) \leq \alpha_1 Q(t_1) + \alpha_2 Q(t_2), \forall\, t_1, t_2 \in (0, 1], \alpha_1, \alpha_2 \geq 0 \text{ s.t. } \alpha_1 + \alpha_2 = 1. \quad \text{(E.2)}$$

For $\forall \alpha_1, \alpha_2 \geq 0$ such that $\alpha_1 + \alpha_2 = 1$, and $t_1, t_2 \in (0, 1)$, we have $\boldsymbol{\Theta}(\alpha_1 t_1 + \alpha_2 t_2) = \alpha_1 \boldsymbol{\Theta}(t_1) + \alpha_2 \boldsymbol{\Theta}(t_2)$. By the bi-linearity property of the inner product function $\langle \cdot, \cdot \rangle$, and using the linearity property of $\boldsymbol{\Theta}(\cdot)$, we have the following equality hold

$$\begin{aligned} &- \big\langle \nabla\mathcal{L}(\boldsymbol{\Theta}^*), \boldsymbol{\Theta}(\alpha_1 t_1 + \alpha_2 t_2) - \boldsymbol{\Theta}^* \big\rangle \\ &= -\alpha_1 \big\langle \nabla\mathcal{L}(\boldsymbol{\Theta}^*), \boldsymbol{\Theta}(t_1) - \boldsymbol{\Theta}^* \big\rangle - \alpha_2 \big\langle \nabla\mathcal{L}(\boldsymbol{\Theta}^*), \boldsymbol{\Theta}(t_2) - \boldsymbol{\Theta}^* \big\rangle. \end{aligned} \quad \text{(E.3)}$$

On the other side, by the convexity of the loss function $\mathcal{L}(\cdot)$, we obtain

$$\mathcal{L}\big(\boldsymbol{\Theta}(\alpha_1 t_1 + \alpha_2 t_2)\big) = \mathcal{L}\big(\alpha_1 \boldsymbol{\Theta}(t_1) + \alpha_2 \boldsymbol{\Theta}(t_2)\big) \leq \alpha_1 \mathcal{L}\big(\boldsymbol{\Theta}(t_1)\big) + \alpha_2 \mathcal{L}\big(\boldsymbol{\Theta}(t_2)\big). \quad \text{(E.4)}$$

By adding (E.3) and (E.4) together and using the definition of function $Q(\cdot)$, we obtain

$$Q(\alpha_1 t_1 + \alpha_2 t_2) \leq \alpha_1 Q(t_1) + \alpha_2 Q(t_2),$$

which indicates $Q(t)$ is a convex function. Thus we complete our proof.

□

**Lemma E.3.** Let $\mathbf{A}_i, \mathbf{B}_i \in \mathbb{R}^{d \times d}$ be square matrices for $i = 1, 2$. Then we have

$$\begin{aligned} \mathbf{A}_1\mathbf{B}_1\mathbf{A}_1 - \mathbf{A}_2\mathbf{B}_2\mathbf{A}_2 = &(\mathbf{A}_1 - \mathbf{A}_2)(\mathbf{B}_1 - \mathbf{B}_2)(\mathbf{A}_1 - \mathbf{A}_2) + (\mathbf{A}_1 - \mathbf{A}_2)\mathbf{B}_2\mathbf{A}_2 \\ &+ (\mathbf{A}_1 - \mathbf{A}_2)\mathbf{B}_2\mathbf{A}_1 + \mathbf{A}_1(\mathbf{B}_1 - \mathbf{B}_2)\mathbf{A}_2. \end{aligned}$$

The next lemma characterizes an upper bound of $\|\mathbf{A}^{-1} - \mathbf{B}^{-1}\|_*$ in terms of $\|\mathbf{A} - \mathbf{B}\|_*$, where $\|\cdot\|_*$ is any matrix norm.

**Lemma E.4.** Let $\mathbf{A}, \mathbf{B} \in \mathbb{R}^{d\times d}$ be invertible. For any matrix norm $\|\cdot\|_*$, we have

$$\|\mathbf{A}^{-1} - \mathbf{B}^{-1}\|_* \le \frac{\|\mathbf{A}^{-1}\|_*^2 \|\mathbf{A} - \mathbf{B}\|_*}{1 - \|\mathbf{A}^{-1}\|_* \|\mathbf{A} - \mathbf{B}\|_*}.$$

We need the following lemma for bounding the difference with respect to the Kronecker product.

**Lemma E.5.** Let $\mathbf{A}$ and $\mathbf{B}$ be matrices of the same dimension. Then we have

$$\|\mathbf{A}\otimes\mathbf{B}\|_\infty = \|\mathbf{A}\|_\infty \|\mathbf{B}\|_\infty, \quad \text{and}$$
$$\|\mathbf{A}\otimes\mathbf{A} - \mathbf{B}\otimes\mathbf{B}\|_\infty \le \|\mathbf{A}-\mathbf{B}\|_\infty^2 + 2\min\big\{\|\mathbf{A}\|_\infty, \|\mathbf{B}\|_\infty\big\}\|\mathbf{A}-\mathbf{B}\|_\infty.$$

The proof of the above lemma can be carried out by using the definitions and thus is omitted here for simplicity.

For a matrix $\mathbf{A} = (a_{ij})$, we say $\mathbf{A}_{\text{sp}} = (a_{ij}^{\text{sp}})$ is the corresponding sparsity pattern matrix if $a_{ij}^{\text{sp}} = 1$ when $a_{ij} \ne 0$; and $a_{ij}^{\text{sp}} = 0$, otherwise.

**Lemma E.6.** Let $\mathbf{A} \in \mathbb{R}^{d\times d}$ be a matrix such that $\|\mathbf{A}\|_{\max} \le 1$. Let $\mathbf{A}_{\text{sp}}$ be the corresponding sparsity pattern matrix. Then we have

$$\|\mathbf{A}\|_2 \le \|\mathbf{A}_{\text{sp}}\|_2.$$

*Proof of Lemma E.6.* Let $a_{ij}$ be the $(i,j)$-th entry of matrix $\mathbf{A}$ and $x_j$ the $j$-th entry of $\mathbf{x}$. Following the definition of the spectral norm of a matrix, we obtain that

$$\begin{aligned}
\|\mathbf{A}\|_2 &= \sup_{\|\mathbf{x}\|_2=1} \|\mathbf{A}\mathbf{x}\|_2 = \sup_{\|\mathbf{x}\|_2=1} \left\{ \sum_{i=1}^n \left( \sum_{j=1}^n a_{ij} x_j \right)^2 \right\} \\
&\le \sup_{\|\mathbf{x}\|_2=1} \left\{ \sum_{i=1}^n \left( \sum_{j=1}^n \operatorname{sgn}(x_j) 1(a_{ij}\ne 0)\cdot x_j \right)^2 \right\} \\
&= \sup_{\mathbf{x}\ge 0, \|\mathbf{x}\|_2=1} \left\{ \sum_{i=1}^n \left( \sum_{j=1}^n 1(a_{ij}\ne 0)\cdot x_j \right)^2 \right\} \le \|\mathbf{A}_{\text{sp}}\|_2.
\end{aligned}$$

Thus the proof is completed. ☐

**Lemma E.7.** Let $\widehat{\mathbf{A}} \in \mathbb{R}^{d\times d}$ be a semi-positive definite random matrix, $\mathbf{A} \in \mathbb{R}^{d\times d}$ a positive definite deterministic matrix. Then we have

$$\mathbb{P}\Big(\big\|\widehat{\mathbf{A}}^{-1} - \mathbf{A}^{-1}\big\|_2 > 2\lambda_{\min}^{-2}(\mathbf{A})\cdot\big\|\widehat{\mathbf{A}} - \mathbf{A}\big\|_2\Big) \le \mathbb{P}\Big(\big\|\widehat{\mathbf{A}} - \mathbf{A}\big\|_2 > 2^{-1}\lambda_{\min}(\mathbf{A})\Big).$$

If we further assume that $\widehat{\mathbf{A}}$ and $\mathbf{A}$ are commutative, that is $\widehat{\mathbf{A}}\mathbf{A} = \mathbf{A}\widehat{\mathbf{A}}$, then we have

$$\begin{aligned}
&\mathbb{P}\Big(\big\|\widehat{\mathbf{A}}^{-1/2} - \mathbf{A}^{-1/2}\big\|_2 > 2(\sqrt{2}+1)\big\|\mathbf{A}\big\|_2^{1/2}\lambda_{\min}^{-2}(\mathbf{A})\big\|\widehat{\mathbf{A}} - \mathbf{A}\big\|_2\Big) \\
&\le \mathbb{P}\Big(\big\|\widehat{\mathbf{A}} - \mathbf{A}\big\|_2 > 2^{-1}\lambda_{\min}(\mathbf{A})\Big).
\end{aligned}$$

*Proof of Lemma E.7.* We first write $\widehat{\mathbf{A}}^{-1}-\mathbf{A}^{-1}$ as $\widehat{\mathbf{A}}^{-1}(\widehat{\mathbf{A}}-\mathbf{A})\mathbf{A}^{-1}$, then it follows from the sub-multiplicative property of the spectral norm that

$$\begin{aligned}\big\|\widehat{\mathbf{A}}^{-1}-\mathbf{A}^{-1}\big\|_2 &\le \big\|\widehat{\mathbf{A}}^{-1}(\widehat{\mathbf{A}}-\mathbf{A})\mathbf{A}^{-1}\big\|_2 \le \big\|\widehat{\mathbf{A}}^{-1}\big\|_2\big\|\mathbf{A}^{-1}\big\|_2\big\|\widehat{\mathbf{A}}-\mathbf{A}\big\|_2\\ &\le \lambda_{\min}^{-1}\big(\widehat{\mathbf{A}}\big)\lambda_{\min}^{-1}\big(\mathbf{A}\big)\cdot\big\|\widehat{\mathbf{A}}-\mathbf{A}\big\|_2. \end{aligned}\tag{E.5}$$

By Weyl's inequality, we obtain that $\lambda_{\min}(\mathbf{A}) \le \lambda_{\min}(\widehat{\mathbf{A}})+\big\|\widehat{\mathbf{A}}-\mathbf{A}\big\|_2$, and thus $\lambda_{\min}\big(\widehat{\mathbf{A}}\big) \ge \lambda_{\min}\big(\mathbf{A}\big) - \big\|\widehat{\mathbf{A}}-\mathbf{A}\big\|_2$. Thus in the event of $\big\{\big\|\widehat{\mathbf{A}}-\mathbf{A}\big\|_2 \le 2^{-1}\lambda_{\min}\big(\mathbf{A}\big)\big\}$, we have $\lambda_{\min}\big(\widehat{\mathbf{A}}\big) \ge 2^{-1}\lambda_{\min}\big(\mathbf{A}\big)$ hold. Thus it follows from (E.5) that

$$\mathbb{P}\Big(\big\|\widehat{\mathbf{A}}^{-1}-\mathbf{A}^{-1}\big\|_2 \le 2\lambda_{\min}^{-2}\big(\mathbf{A}\big)\cdot\big\|\widehat{\mathbf{A}}-\mathbf{A}\big\|_2\Big) \ge \mathbb{P}\Big(\big\|\widehat{\mathbf{A}}-\mathbf{A}\big\|_2 \le 2^{-1}\lambda_{\min}\big(\mathbf{A}\big)\Big).$$

This proves the first desired probability bound. If we further assume that $\widehat{\mathbf{A}}$ and $\mathbf{A}$ are commutative, under the event $\big\{\big\|\widehat{\mathbf{A}}-\mathbf{A}\big\|_2 \le 2^{-1}\lambda_{\min}\big(\mathbf{A}\big)\big\}$, we have

$$\begin{aligned}\big\|\widehat{\mathbf{A}}^{-1/2}-\mathbf{A}^{-1/2}\big\|_2 &= \big\|\big(\widehat{\mathbf{A}}^{-1/2}+\mathbf{A}^{-1/2}\big)^{-1}\big(\widehat{\mathbf{A}}^{-1}-\mathbf{A}^{-1}\big)\big\|_2\\ &\le \Big(\big\|\widehat{\mathbf{A}}\big\|_2^{1/2}+\big\|\mathbf{A}\big\|_2^{1/2}\Big)\big\|\widehat{\mathbf{A}}^{-1}-\mathbf{A}^{-1}\big\|_2\\ &\le (\sqrt{2}+1)\big\|\mathbf{A}\big\|_2^{1/2}\big\|\widehat{\mathbf{A}}^{-1}-\mathbf{A}^{-1}\big\|_2\\ &\le 2(\sqrt{2}+1)\big\|\mathbf{A}\big\|_2^{1/2}\lambda_{\min}^{-2}\big(\mathbf{A}\big)\big\|\widehat{\mathbf{A}}-\mathbf{A}\big\|_2.\end{aligned}$$

Therefore we prove the third result.

□

The following lemma is taken from Dembo and Zeitouni (2009), which leads to a concentration bound of the empirical means $\bar{X} = n^{-1}\sum_{i=1}^n X_i$, where $X_i$'s are i.i.d. random copies of $X$. Define the logarithmic moment generating function associated with $X$ to be

$$\Lambda_X(\lambda) \equiv \log M_X(\lambda) = \log\mathbb{E}\big[\exp\{\lambda X\}\big]. \tag{E.6}$$

**Lemma E.8** (Large Deviation Inequality)**.** Let the logarithmic moment generating function of $X$, $\Lambda_X(\lambda)$, be defined in E.6. Define the Fenchel-Legendre dual of $\Lambda_X(x)$ to be $\Lambda_X^*(x) \equiv \sup_{\lambda\in\mathbb{R}}\big\{\lambda x - \Lambda(\lambda)\big\}$. Then, for any $t \ge 0$, we have

$$\mathbb{P}\bigg(\frac{1}{n}\sum_{i=1}^n X_i - \mathbb{E}X \ge t\bigg) \le \exp\big\{-n(\mathbb{E}X + \inf_{x\in F_1}\Lambda^*(x))\big\} \ \text{ and}$$
$$\mathbb{P}\bigg(\frac{1}{n}\sum_{i=1}^n X_i - \mathbb{E}X \le -t\bigg) \le \exp\big\{-n(\mathbb{E}X + \inf_{x\in F_2}\Lambda^*(x))\big\},$$

where $F_1 = \big[t,+\infty\big)$ and $F_2 = \big(-\infty,-t\big]$.